\pdfoutput=1

\documentclass[11pt]{article}
\RequirePackage[OT1]{fontenc}
\RequirePackage{amsthm,amsmath}
\RequirePackage[numbers]{natbib}
\RequirePackage[colorlinks,citecolor=blue,urlcolor=blue]{hyperref}

\usepackage[T1]{fontenc}    
\usepackage[colorlinks]{hyperref} 
\usepackage{url}            
\usepackage{booktabs}       
\usepackage{amsfonts}       
\usepackage{amsmath}
\usepackage{nicefrac}       
\usepackage{microtype}      
\usepackage{color}
\usepackage{bbm}
\usepackage{amssymb, enumerate}
\usepackage{makecell}
\usepackage{wrapfig}
\usepackage{extarrows}
\usepackage{arydshln}

\usepackage{amsmath,amsfonts,amsthm, amssymb, enumerate}
\usepackage{lipsum}
\usepackage{caption}
\usepackage{amssymb,graphicx,subfigure}
\usepackage[bottom]{footmisc}
\usepackage[dvipsnames]{xcolor}
\usepackage{array,multirow}
\usepackage{enumitem}

\usepackage[font=small,labelfont=bf]{caption}

\newtheorem{defn}{Definition}

\newtheorem{thm}{Theorem}

\newcommand{\argmin}{\operatornamewithlimits{argmin}}
\newcommand{\argmax}{\operatornamewithlimits{argmax}}

\usepackage{algorithm,algpseudocode}
\algnewcommand\TR{\item[{\textbf{Training phase}}]}
\algnewcommand\TE{\item[{\textbf{Test phase}}]}
\algnewcommand\Input{\item[{{Input:}}]}
\algnewcommand\Output{\item[{{Output:}}]}
\algnewcommand\Initialize{\item[{{Initialize:}}]}
\algnewcommand{\return}[1]{
	\State \textbf{return:}
	\Statex \hspace*{\algorithmicindent}\parbox[t]{.8\linewidth}{\raggedright #1}
}
\makeatletter
\newcommand{\algrule}[1][.2pt]{\par\vskip.5\baselineskip\hrule height #1\par\vskip.5\baselineskip}
\makeatother

\usepackage{setspace}
\begin{document}
	\title{Compressed Sensing via Measurement-Conditional Generative Models}
	\date{}
	\author{
		Kyung-Su Kim$^{1}$\thanks{Equal contribution}, Jung Hyun Lee$^{2}$\footnotemark[1], Eunho Yang$^{2}$\\
		{\small kskim.doc@gmail.com, junghyunlee@kaist.ac.kr, eunhoy@kaist.ac.kr}\\
		{\small $^{1}$AI Research Institute, Samsung Seoul Medical Center, Seoul, South Korea} \\
		{\small $^{2}$Korea Advanced Institute of Science and Technology (KAIST), Daejeon, South Korea}
	} 
	\maketitle 
	
	\begin{abstract}
    A pre-trained generator has been frequently adopted in compressed sensing (CS) due to its ability to effectively estimate signals with the prior of NNs. In order to further refine the NN-based prior, we propose a framework that allows the generator to utilize additional information from a given measurement for prior learning, thereby yielding more accurate prediction for signals. As our framework has a simple form, it is easily applied to existing CS methods using pre-trained generators. We demonstrate through extensive experiments that our framework exhibits uniformly superior performances by large margin and can reduce the reconstruction error up to an order of magnitude for some applications. We also explain the experimental success in theory by showing that our framework can slightly relax the stringent signal presence condition, which is required to guarantee the success of signal recovery.
	\end{abstract}
	
\section{Introduction}
Compressed Sensing (CS) has been a popular approach for decades to recover signals when the number of devices is larger than the size of measurements like in communications \citep{he2018deep,kim2019tree} or measurements are expensive such as in medical imaging \citep{lustig07csmri,sun2016deep} 
and optical imaging \citep{willett2011compressed}. CS aims to estimate a signal $\boldsymbol{x}  \in \mathbb{R}^{d}$ given an undersampled measurement vector $\boldsymbol{y} \in \mathbb{R}^{m}$ under the following linear relationship:
\begin{align}\label{CS}
\boldsymbol{y} =\boldsymbol{A}\boldsymbol{x} + \boldsymbol{\omega}, 
\end{align} 
where $\boldsymbol{A} \in \mathbb{R}^{m \times d}$ is a given sensing matrix such that $m < d$, and $\boldsymbol{\omega}$ is a unknown noise. Since \eqref{CS} is an underdetermined linear system, it requires some underlying assumption about the signal  to guarantee a unique solution. Classical literature on CS postulates that $\boldsymbol{x}$ would be sparse in some known basis and solves \eqref{CS} by $\ell_1$-minimization.

As neural networks (NNs) have accomplished enormous success in both supervised learning including regression and classification tasks and unsupervised learning such as clustering and density estimation tasks, many researchers have recently devoted much effort to leveraging NNs as a structural assumption for CS \citep{sun2016deep,bora2017compressed,he2017bayesian,metzler2017learned,van2018compressed,dhar2018modeling,mardani2018neural,lunz2018adversarial,mousavi2018data,grover2019uncertainty,wu2019learning,wu2019deep,raj2019gan}. 
In particular, \citet{bora2017compressed} proves that CS using pre-trained generators (CSPG) is able to reconstruct signals by taking advantage of a domain-specific prior instead of sparsity prior. 
Although methods using LASSO \citep{tibshirani1996lasso} only capture signal sparsity from data transformed by a certain operator (e.g., the wavelet transform), real-world data possess a variety of features other than the sparsity. As a result, training a generative model enables its generator to learn a domain-specific distribution, which allows for signal recovery even with a fewer number of measurements than theoretical lower bounds under sparsity prior. 

Unfortunately, prior works \citep{bora2017compressed, dhar2018modeling, shah2018solving, kabkab2018task, wu2019deep, grover2019uncertainty, raj2019gan} in CSPG solely concentrate on training generators without conditioning on any measurement vector. As $\boldsymbol{y}$ can be viewed as a (linearly) compressed signal of the original one $\boldsymbol{x}$, it includes the information of $\boldsymbol{x}$, thus helping generators trained to estimate $\boldsymbol{x}$ by additionally using $\boldsymbol{y}$. Motivated by this, we make a generator take a measurement vector as additional input to learn the signal more effectively.

Our contribution is threefold: 
\vspace{-0.25cm}
\begin{itemize}[leftmargin=8mm]
    \item We propose a simple but effective framework allowing a generator to exploit $\boldsymbol{y}$ directly in learning its parameters to better estimate its true signal. To the best of our knowledge, our framework is the first attempt to insert $\boldsymbol{y}$ into generative models for CS.
    \item We provide supporting theoretical insight that our framework alleviates the stringent signal presence assumption, thus making signal reconstruction much more successful. 
    \item We empirically show consistent and considerable improvements on a wide variety of prior works.  
    We further demonstrate the practicability of our method on real-world data that are difficult for previous generative models to reconstruct  (i.e., MRI image reconstruction).
\end{itemize}
\vspace{-0.3cm}
 
\section{Related Work} 

We introduce several lines of research in CS using NNs, which can be largely split into two groups relying on whether to make use of generators or not.

\vspace{-0.2cm}
\paragraph{CS via NNs without generators} The first group is concerned with devising NN architectures for special purposes. 
\citet{gregor2010learning} suggested that the update step in the iterative shrinkage-thresholding algorithm (ISTA) could be represented as each layer of a NN and proposed a deep architecture as a learned variant of ISTA (LISTA). As LISTA optimizes the network whose form is initially set to ISTA, LISTA directly improves the performance of ISTA by using its architecture. Motivated by this unfolding procedure, extensive studies \citep{moreau2017understanding,giryes2018tradeoffs,chen2018theoretical,tramel2016approximate,borgerding2017amp,metzler2017learned,he2017bayesian,sun2016deep, mardani2018neural} have been conducted by unfolding state-of-the-art CS algorithms (e.g., approximate message passing, sparse Bayesian learning, and alternating direction method of multipliers) and mapping them to certain network structures. In addition to unfolding-based research, \citet{mousavi2018data} proposed a variant of a convolutional autoencoder to accelerate signal recovery and induce a data-driven dimensionality reduction. \citet{wu2019learning} presented how to learn a sensing matrix via designing an autoencoder inspired by the projected subgradient method. \citet{lunz2018adversarial} studied the case where the regularization functional is built as a NN.

\vspace{-0.2cm}
\paragraph{CS via NNs with generators} The other group is further classified into CS using pre-trained generators (CSPG) and CS using untrained generators (CSUG) depending on whether to train a generator or not. CSPG indicates algorithms to recover signals by the aid of generators trained over data. \citet{bora2017compressed} first employed pre-trained generators to reconstruct signals, providing a recovery guarantee. A number of studies \citep{dhar2018modeling,shah2018solving,raj2019gan,kabkab2018task,grover2019uncertainty, wu2019deep} have been done to enhance the performance of CSPG thereafter. In contrast to CSPG, CSUG \citep{van2018compressed, jagatap2019algorithmic} represents methods based on the deep image prior \citep{ulyanov2018dip} so that the weights of an untrained generator can be trained by using only one measurement vector $\boldsymbol{y}_{te}$. Although CSUG is able to recover a signal even in the absence of training data, CSPG is more widely used than CSUG because training data can be commonly collected in practice and CSPG usually outperforms CSUG (e.g., CSUG \citep{heckel2018deep, jagatap2019algorithmic} performs similarly to a wavelet-based LASSO method, but CSPG \citep{bora2017compressed,dhar2018modeling} outperforms it).

In this work, we propose a new framework that is easily applicable to existing CSPG methods while significantly and uniformly improving them by modeling conditional generative models like conditional GAN (cGAN) \citep{mirza2014conditional}. 
Although cGAN is applied to various fields \citep{reed2016textgan,isola2017image,wang2018high,uelwer2019prcgan,ye2020deep}, our paper first delve into the application of cGAN to CS.  


\section{Measurement-Conditional Generative Models for CS} 
In this section, we outline an essential background of CSPG and propose a simple yet effective scheme to cope with this limitation.


\vspace{-0.15cm}
\paragraph{Notation} $G_{\boldsymbol{\theta}}(\boldsymbol{z})$ denotes a generator with parameters $\boldsymbol{\theta}$ and latent variables $\boldsymbol{z}$.  $D_{\boldsymbol{\phi}}(\bar{\boldsymbol{x}})$ re-presents a discriminator with parameters $\phi$ and input $\bar{\boldsymbol{x}}$. $(\boldsymbol{x}_{tr}, \boldsymbol{y}_{tr})$ (or $(\boldsymbol{x}_{te}, \boldsymbol{y}_{te})$) indicates a pair of a training (or test) signal $\boldsymbol{x}_{tr}$ (or $\boldsymbol{x}_{te}$) and the corresponding measurement vector $\boldsymbol{y}_{tr}$ (or $\boldsymbol{y}_{te}$) by \eqref{CS}. $\textup{supp}(\boldsymbol{x})$ denotes the support of $\boldsymbol{x}$.

\subsection{Preliminary: compressed sensing using pre-trained generators}\label{subsec:preliminary}

Algorithms in CS using pre-trained generators (CSPG)\footnote{Most CSPG methods are based on Generative Adversarial Networks (GANs) \citep{goodfellow2014gan, radford2015dcgan}. Hence, we focus on GAN-based prior generators throughout the paper, but it can be seamlessly extended to other generative models such as Variational Autoencoder \citep{kingma2014vae}} can be typically divided into the following two phases, training \eqref{step1}  and test \eqref{step2}, respectively. 

The training phase aims to find optimal parameters $(\boldsymbol{\theta}^*, \boldsymbol{\phi}^*)$ of $G_{\boldsymbol{\theta}}$ and $D_{\boldsymbol{\phi}}$ given training signals:
\begin{align}\label{step1}
    (\boldsymbol{\theta}^*, \boldsymbol{\phi}^*)= \mathcal{F}^{opt}_{(\boldsymbol{\theta},\boldsymbol{\phi})}\Big[\mathcal{L}_{tr}\big(G_{\boldsymbol{\theta}}(\boldsymbol{z}), D_{\boldsymbol{\phi}}(\bar{\boldsymbol{x}})\big)\Big],
\end{align}
where $\bar{\boldsymbol{x}}$ denotes either a training signal $\boldsymbol{x}_{tr}$ or a \emph{fake} signal generated by $G_{\boldsymbol{\theta}}(\boldsymbol{z})$, $\mathcal{L}_{tr}(\mathcal{M})$ indicates a loss function to train models $\mathcal{M}$, and $\mathcal{F}^{opt}_{\mathcal{T}}[\cdot]$ is defined by an operator to optimize its input with respect to $\mathcal{T}$. In general, $\mathcal{M} = \{G_{\boldsymbol{\theta}}(\boldsymbol{x}), D_{\boldsymbol{\phi}}(\bar{\boldsymbol{x}})\}$ and $\mathcal{T} = \{\boldsymbol{\theta}, \boldsymbol{\phi}\}$, but $\mathcal{T}$ can vary depending on $\mathcal{M}$.

In the test phase, a target signal $\boldsymbol{x}_{te}$ is estimated as $\hat{\boldsymbol{x}}$ by the following two-stage process: we first find the optimal latent variables $\boldsymbol{z}^*$ given trained $\boldsymbol{\theta}^*$ and measurement $\boldsymbol{y}_{te}$:
\begin{align}\label{step2}
    \boldsymbol{z}^*  = \mathcal{F}^{opt}_{\boldsymbol{z}}\Big[\mathcal{L}_{te}\big(G_{\boldsymbol{\theta}^*}(\boldsymbol{z}) , \boldsymbol{y}_{te}\big)\Big] , 
\end{align} 
where $\mathcal{L}_{te}\big(G_{\boldsymbol{\theta}^*}(\boldsymbol{z}) ,  \boldsymbol{y}_{te}\big)$ represents an objective function for inference. Then, given estimated $\boldsymbol{z}^*$ and trained $\boldsymbol{\theta}^*$, we recover the target signal $\boldsymbol{x}_{te}$ as  $\hat{\boldsymbol{x}} = G_{\boldsymbol{\theta}^*}(\boldsymbol{z}^*)$.


\subsection{Measurement-Conditional Pre-Trained Generators}\label{subsec:framework}

\begin{wrapfigure}{r}{0.4\textwidth}
	\centering
	\vspace{-0.25cm}
	\subfigure{\includegraphics[width=1\linewidth]{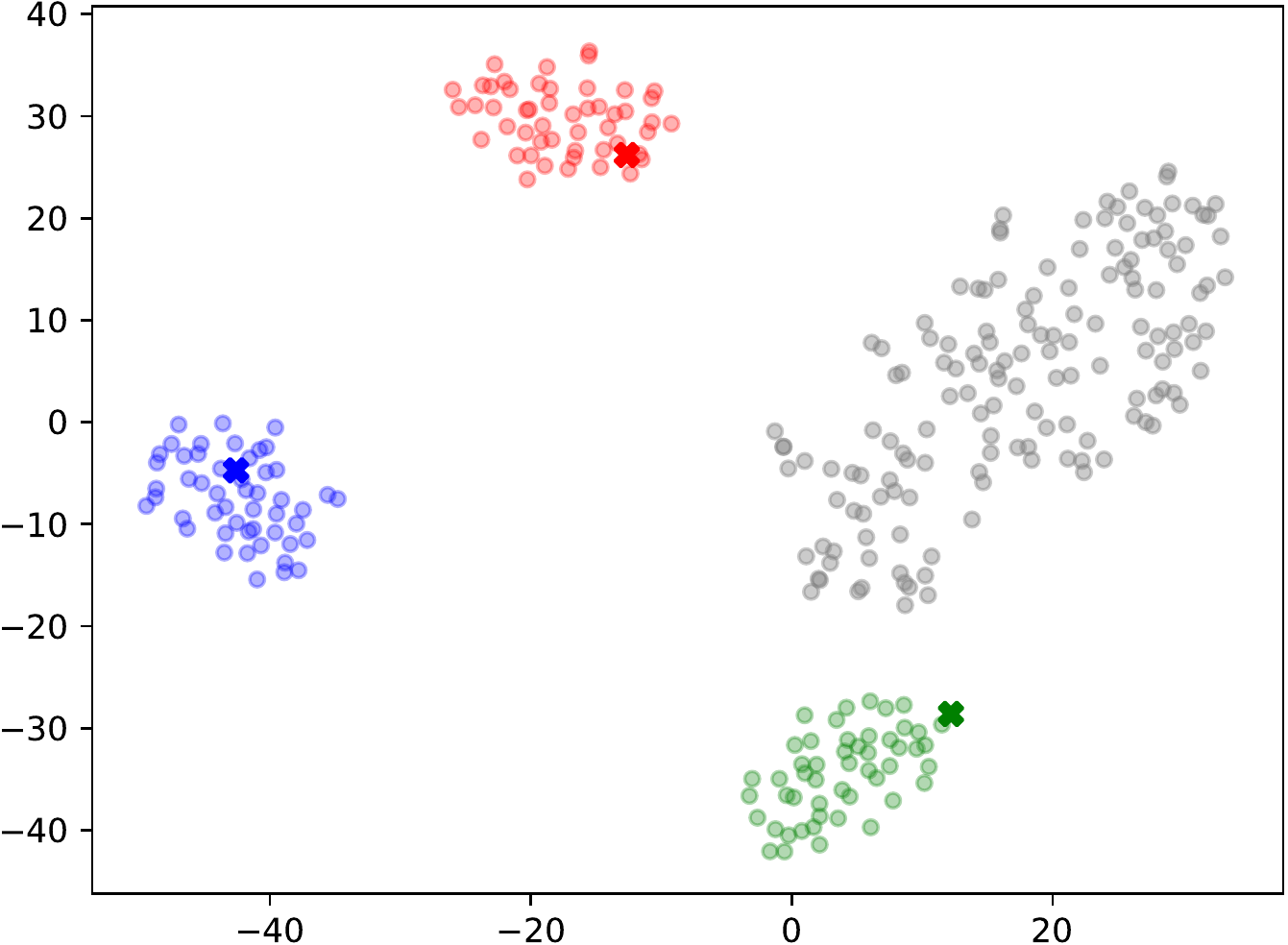}} 
	\caption{\footnotesize t-SNE visualization \citep{vandermaaten2008tsne} of samples from a marginal generator and its conditional counterpart, respectively. Gray points come from a marginal generator and red/blue/green-colored points are samples from a conditional generator on each colored `x'. Here, $m=1000$.
	}
	\label{fig:im_effect}
	\vspace{-0.3cm}
\end{wrapfigure}

In contrast to existing CSPG methods such as DCGAN \citep{radford2015dcgan}, our framework provides a way to exploit the measurement information in learning the generative model by leveraging the concept of conditional GAN \citep{mirza2014conditional}. Essentially, we generate a prior distribution of signals not just by `noise', but also with specific `measurement information', thus further refining the prior with the additional information of measurement vectors. Figure \ref{fig:im_effect} illustrates that each colored points are much closer to its target signal marked by `x' than gray points, which implies that a measurement-conditional generator is able to provide a more refined prior for each target signal than its marginal counterpart, thereby increasing the chance of finding $\boldsymbol{z}^*$ that estimates a target signal $\boldsymbol{x}_{te}$. 

Overall,  
we inject the measurement information into $G_{\boldsymbol{\theta}}$ and $D_{\boldsymbol{\phi}}$ as an input for both the training and test phases so that \eqref{step1} and \eqref{step2} are modified to \eqref{imgeneral1} and \eqref{imgeneral2} respectively:  
\begin{align}
     &(\boldsymbol{\theta}^*, \boldsymbol{\phi}^*) = \mathcal{F}^{opt}_{(\boldsymbol{\theta},\boldsymbol{\phi})}\Big[\mathcal{L}_{tr}\big(G_{\boldsymbol{\theta}}(\boldsymbol{z},\textcolor{black}{\boldsymbol{y}_{tr}}), D_{\boldsymbol{\phi}}(\bar{\boldsymbol{x}},\textcolor{black}{\boldsymbol{y}_{tr}})\big)\Big] \label{imgeneral1} \\ 
     &\boldsymbol{z}^* = \mathcal{F}^{opt}_{\boldsymbol{z}}\Big[\mathcal{L}_{te}\big(G_{\boldsymbol{\theta}^*}(\boldsymbol{z},\textcolor{black}{\boldsymbol{y}_{te}}) \, , \, \boldsymbol{y}_{te}\big)\Big], \,\, \hat{\boldsymbol{x}} = G_{\boldsymbol{\theta}^*}(\boldsymbol{z}^*,\textcolor{black}{\boldsymbol{y}_{te}}) \label{imgeneral2}  
\end{align}

This approach, coined `Inserting Measurements' (IM), is readily applicable to any CS method using pre-trained generators (CSPG) as shown in Figure \ref{fig:csgm_and_im}.

Unlike discriminative models (i.e., learning NNs for CS that directly map measurements to target signals without any generator), a generative model for CS has latent variables $\boldsymbol{z}$, which can be optimized in the test phase via \eqref{step2} to find an estimate closer to the true signal. 
Our method, IM, also make full use of such a latent optimization as demonstrated in Figure \ref{fig:average_reconstruction_errors} and \ref{before vs. after}. Through the latent optimization in \eqref{imgeneral2}, not only does the reconstruction error of IM decrease continuously as the latent optimization progresses, but also reconstructed images after optimizing $z$ more resemble the original images than those before optimizing $z$.
Hence, we can confirm that our measurement-conditional generative model inherits both advantages of generative models (i.e., latent optimization) and discriminative models (i.e., directly taking $\boldsymbol{y}$ as input to estimate $\boldsymbol{x}$). 
\section{Revising Existing CSPG Models under Our Framework}
In this section, we delineate how prior studies such as Compressed Sensing using Generative Models (CSGM) \citep{bora2017compressed} and Projected Gradient Descent GAN (PGDGAN) \citep{shah2018solving} are modified under our framework, IM. The applications of IM to Deep Compressed Sensing (DCS) \citep{wu2019deep} and SparseGen \citep{dhar2018modeling} are deferred to Appendix.

\subsection{Compressed Sensing using Generative Models (CSGM)}\label{subsec:csgm}

\begin{figure*}[t]
	\centering
	\subfigure{\includegraphics[width=1\linewidth]{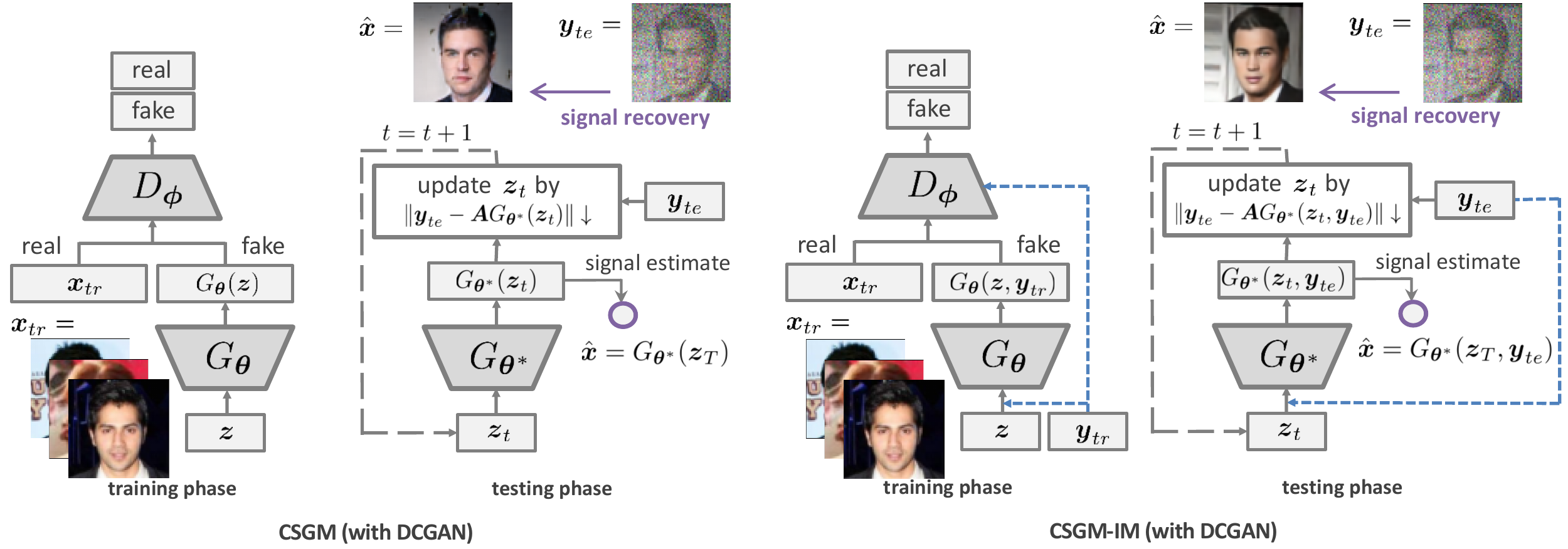}}
	\caption{\footnotesize Illustration of CSGM and CSGM-IM with the DCGAN architecture.} 
	\label{fig:csgm_and_im}
\end{figure*}

Compressed Sensing using Generative Models (CSGM) \citep{bora2017compressed} is the first work to propose the GAN-based CSPG framework in \eqref{step1} and \eqref{step2}. While it permits any GAN models, if the standard GAN training objective including DCGAN \citep{radford2015dcgan} is used as showed in \citet{bora2017compressed}, 
a generator $G_{\boldsymbol{\theta}}$ and a discriminator $D_{\boldsymbol{\phi}}$ are trained by solving the following min-max problem: 
\begin{align}\label{dcgan_obj} 
	    (\theta^*,\phi^*) \leftarrow  \argmin_{\boldsymbol{\theta}}&\argmax_{\boldsymbol{\phi}}\, 
	    \mathbb{E}_{\boldsymbol{x}_{tr} \sim p(\boldsymbol{x})}\Big[\ln \, D_{\boldsymbol{\phi}}(\boldsymbol{x}_{tr})\Big] + \mathbb{E}_{\boldsymbol{z} \sim p_{\boldsymbol{z} }(\boldsymbol{z})}\Big[\ln \Big(1-D_{\boldsymbol{\phi}}\big( G_{\boldsymbol{\theta}}(\boldsymbol{z})\big)\Big)\Big].
\end{align} 
Another reasonable GAN instance under its framework is BEGAN. In this case, the training phase of CSGM can be given as
\begin{align}\label{began_obj} 
&(\theta^*,\phi^*) \leftarrow \argmax_{\boldsymbol{\theta}}\argmin_{\boldsymbol{\phi}}\, 
\mathbb{E}_{\boldsymbol{x}_{tr} \sim p(\boldsymbol{x}),\boldsymbol{z} \sim p_{\boldsymbol{z}}}\Big[ R_{\boldsymbol{\phi}}\big(\boldsymbol{x}_{tr}\big) - \zeta R_{\boldsymbol{\phi}}\big(G_{\boldsymbol{\theta}}(\boldsymbol{z})\big)\Big], 
\end{align}
where $R_{\boldsymbol{\phi}}(\bar{\boldsymbol{x}})=\left\|\bar{\boldsymbol{x}}- D_{\boldsymbol{\phi}}(\bar{\boldsymbol{x}})\right\|_p$ denotes the recon-struction loss determined by $D_{\boldsymbol{\phi}}$ in terms of $p = 1 \text{ or } 2$ norms and the parameter $\zeta$ controls the balance between auto-encoding true images and distinguishing real images from fake ones. While $D_{\phi}$ in \eqref{dcgan_obj} outputs a scalar value (probability) indicating whether its input is a real image or not, $D_{\phi}$ in \eqref{began_obj} reconstructs an image based on its input.

CSGM in the test phase computes an estimate $\hat{\boldsymbol{x}}=G_{\boldsymbol{\theta^*}}(\boldsymbol{z}^*)$ of target signal $\boldsymbol{x}_{te}$ by minimizing the following loss with respect to the input noise $\boldsymbol{z}$ of the pre-trained generator $G_{\boldsymbol{\theta^*}}$:
\begin{align}\label{csgm} 
	   \boldsymbol{z}^* \leftarrow \argmin_{\boldsymbol{z}} \left\| \boldsymbol{y}_{te} - \boldsymbol{A}G_{\boldsymbol{\theta^*}}(\boldsymbol{z}) \right\|^2,\,\,\,\,\, \hat{\boldsymbol{x}} = G_{\boldsymbol{\theta}^*}(\boldsymbol{z}^*).
\end{align}

\vspace{-0.1cm}
\paragraph{CSGM-IM}
We show how our framework, IM can be applied to CSGM, naming it CSGM-IM.  

The training phase of CSGM-IM learns  $(\boldsymbol{\theta},\boldsymbol{\phi})$ according to the following optimization \eqref{im_dcgan_obj} (for DCGAN) and \eqref{im_began_obj} (for BEGAN) respectively: 
\begin{align}
    (\theta^*,\phi^*)
    \leftarrow &\argmin_{\boldsymbol{\theta}}\argmax_{\boldsymbol{\phi}}\, \mathbb{E}_{\boldsymbol{x}_{tr} \sim p(\boldsymbol{x})}\Big[\ln \, D_{\boldsymbol{\phi}}(\boldsymbol{x}_{tr}, \begin{color}{black}\boldsymbol{y}_{tr}\end{color})\Big] + \mathbb{E}_{\boldsymbol{z} \sim p_{\boldsymbol{z} }(\boldsymbol{z})}\Big[\ln \, \Big(1-D_{\boldsymbol{\phi}}\big( G_{\boldsymbol{\theta}}(\boldsymbol{z}, \begin{color}{black}\boldsymbol{y}_{tr}\end{color}), \begin{color}{black}\boldsymbol{y}_{tr}\end{color}\big)\Big)\Big]\label{im_dcgan_obj} 
\end{align}
\vspace{-0.5cm}
\begin{align}
    (\theta^*,\phi^*) 
    \leftarrow \argmax_{\boldsymbol{\theta}}\argmin_{\boldsymbol{\phi}}&\, \mathbb{E}_{\boldsymbol{x}_{tr} \sim p(\boldsymbol{x}),\boldsymbol{z} \sim p_{\boldsymbol{z}}}\Big[ R_{\boldsymbol{\phi}}(\boldsymbol{x}_{tr}, \textcolor{black}{\boldsymbol{y}_{tr}})- \zeta R_{\boldsymbol{\phi}}\big(G_{\boldsymbol{\theta}}(\boldsymbol{z}, \textcolor{black}{\boldsymbol{y}_{tr}}), \boldsymbol{y}_{tr}\big)\Big] \label{im_began_obj} 
\end{align} 
where $R_{\boldsymbol{\phi}}(\bar{\boldsymbol{x}},\boldsymbol{y})=\left\|\bar{\boldsymbol{x}}- D_{\boldsymbol{\phi}}(\bar{\boldsymbol{x}},\boldsymbol{y})\right\|_p$ represents the reconstruction loss determined by the modified discriminator $D_{\boldsymbol{\phi}}(\bar{\boldsymbol{x}}, \boldsymbol{y})$.

In the test phase, CSGM-IM estimates the target signal $\boldsymbol{x}_{te}$ as $\hat{\boldsymbol{x}} = G_{\boldsymbol{\theta^*}}(\boldsymbol{z}^*, \boldsymbol{y}_{te})$ by consistently feeding the test measurement to the learned generator:
\begin{align}\label{csgm_im} 
	   \boldsymbol{z}^* \leftarrow 
	   \underset{\boldsymbol{z}}{\arg \min} \left\| \boldsymbol{y}_{te} - \boldsymbol{A}G_{\boldsymbol{\theta^*}}(\boldsymbol{z},\textcolor{black}{\boldsymbol{y}_{te}}) \right\|^2,
	   \,\,
	   \hat{\boldsymbol{x}} = G_{\boldsymbol{\theta}^*}(\boldsymbol{z}^*,\textcolor{black}{\boldsymbol{y}_{te}}).
\end{align} 
We illustrate how CSGM-IM differs from CSGM in Figure \ref{fig:csgm_and_im} and defer their algorithms to Appendix. 


\begin{figure*}
    \centering
    \subfigure[\footnotesize Test phases of an existing scheme and its IM \label{fig:average_reconstruction_errors}]{\includegraphics[width=0.45\linewidth ]{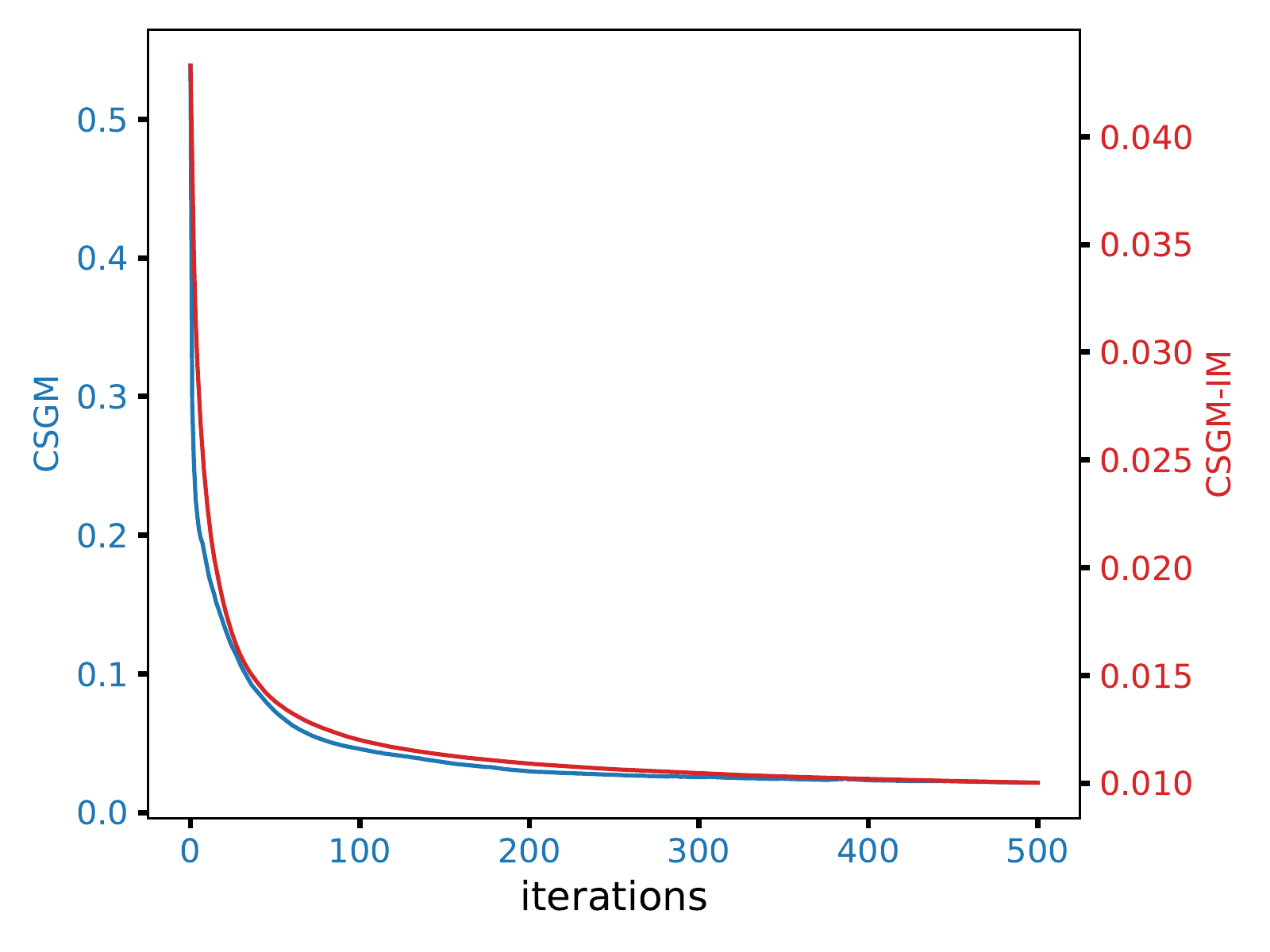}} 
    \subfigure[\footnotesize Before vs. After optimizing $z$ \label{before vs. after}]{\includegraphics[width=0.45\linewidth, height=5.5cm ]{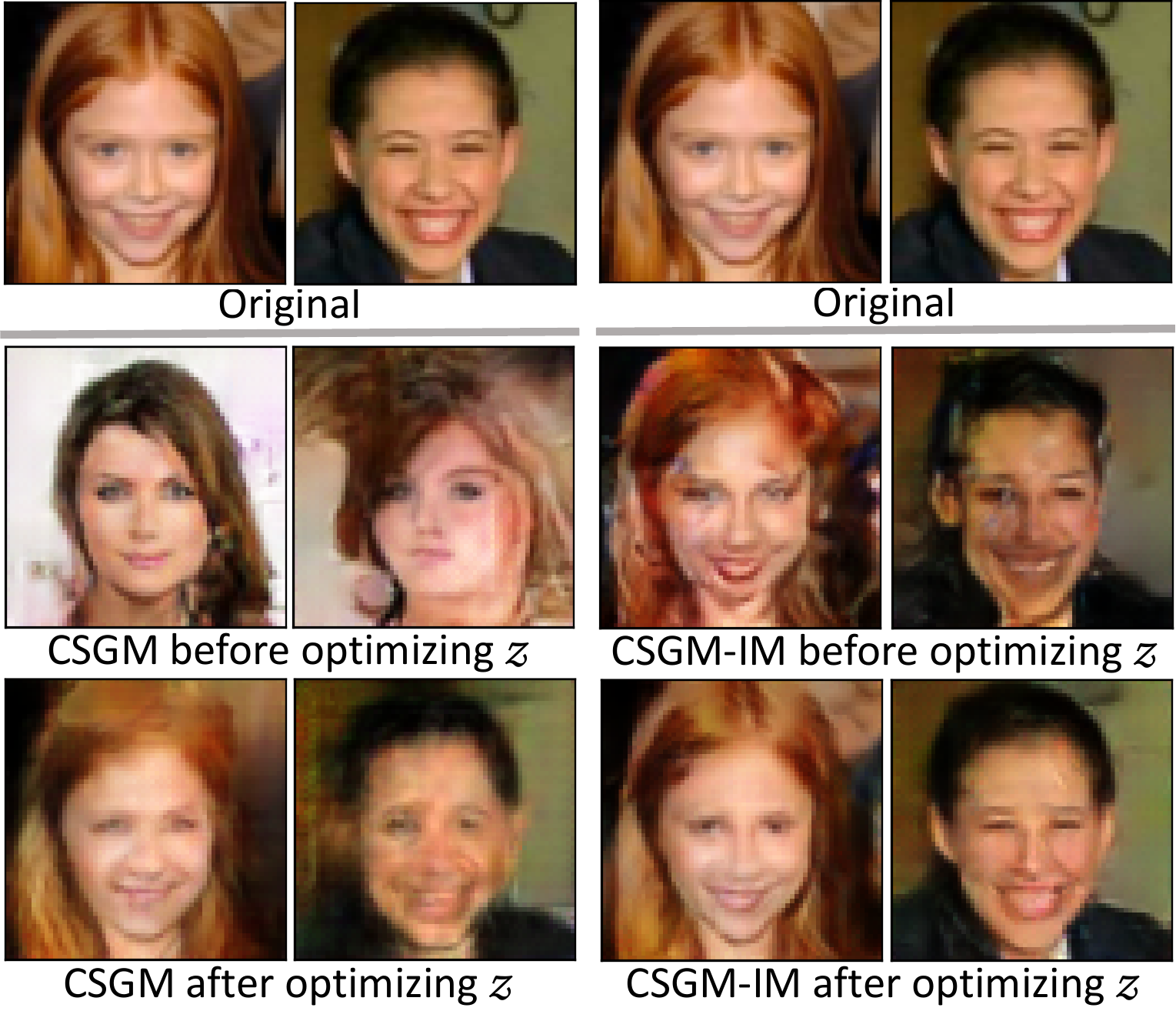}} 
    \vskip -8pt
    \caption{\footnotesize 
    In Figure \ref{fig:average_reconstruction_errors}, the blue and red curves correspond to the reconstruction error of an existing approach using a marginal generator and that of our proposed approach using a measurement-conditional generator, respectively. Figure \ref{before vs. after} shows recovered images on CelebA before and after optimizing $z$. Here, $m=1000$. 
    } 
    \label{fig:z_effect}
\end{figure*}
\subsection{Projected Gradient Descent GAN (PGDGAN) and Its Sparse Variant}\label{subsec:pgd}
Projected Gradient Descent GAN (PGDGAN) \citep{shah2018solving} is a representative work of adopting projected gradient descent under CSPG. Similarly to CSGM, the training phase of PGDGAN can be any learning scheme to optimize a generator $G_{\boldsymbol{\theta}}$ like \eqref{dcgan_obj} or \eqref{began_obj}. 
PGDGAN in the test phase, however, computes an estimate $\hat{\boldsymbol{x}}=\boldsymbol{x}_{T}$ of the target signal $\boldsymbol{x}_{te}$ by iteratively solving the following recursive formula with respect to the input noise $\boldsymbol{z}$ of a pre-trained generator $G_{\boldsymbol{\theta^*}}$:  
\begin{align}\label{pgd_test}
  &\boldsymbol{x}_{t+1} = \mathcal{P}_{G_{\boldsymbol{\theta}^*}}\Big(\boldsymbol{x}_{t} - \alpha   \frac{1}{2} \nabla \left\| \boldsymbol{y}_{te} - \boldsymbol{A} \boldsymbol{x}_{t}   \right\|^2\Big) = G_{\boldsymbol{\theta}^*}\Big(\underset{\boldsymbol{z}}{\arg\min}\left\| \boldsymbol{x}_{t} - \alpha  \boldsymbol{A}^{\top}(\boldsymbol{A} \boldsymbol{x}_{t} - \boldsymbol{y}_{te}) - G_{\boldsymbol{\theta}^*}(\boldsymbol{z})\right\|\Big),
\end{align}
for $t = 0, \cdots, T-1$, where $\mathcal{P}_{G_{\boldsymbol{\theta}^*}}(\boldsymbol{h})=G_{\boldsymbol{\theta}^*}(\argmin_{\boldsymbol{z}}\left\|\boldsymbol{h} - G_{\boldsymbol{\theta}^*}(\boldsymbol{z})\right\|)$ denotes an operator projecting the input $\boldsymbol{h}$ onto the range of the pre-trained generator $G_{\boldsymbol{\theta}^*}$, $\boldsymbol{x}_0 = \boldsymbol{0} \text{ or } \boldsymbol{A}^{\top}\boldsymbol{y}_{te}$ in general, $\alpha$ is a learning rate, and $T$ is the total number of iterations. 

\vspace{-0.1cm}

\paragraph{PGDGAN-IM}
We describe how IM is applicable to PGDGAN, which is dubbed PGDGAN-IM.

As PGDGAN has the same training phase as CSGM, PGDGAN-IM also has the same training phase as CSGM-IM (\eqref{im_dcgan_obj} or \eqref{im_began_obj}). In the test phase, PGDGAN-IM estimates the target signal $\boldsymbol{x}_{te}$ as  $\hat{\boldsymbol{x}}=\boldsymbol{x}_{T}$ by consistently feeding the test measurement to the trained generator:
\begin{align}\label{pgdim_test}
	\boldsymbol{z}_{t+1} &= \underset{\boldsymbol{z}}{\arg\min}\left\| \boldsymbol{x}_{t} - \alpha  \boldsymbol{A}^{\top}(\boldsymbol{A} \boldsymbol{x}_{t} - \boldsymbol{y}_{te}) - G_{\boldsymbol{\theta}^*}(\boldsymbol{z},\boldsymbol{y}_{te})\right\|, \nonumber \\
	\boldsymbol{x}_{t+1} &= G_{\boldsymbol{\theta}^*}(\boldsymbol{z}_{t+1},\boldsymbol{y}_{te}).
\end{align}
We provide the algorithms of PGDGAN and PGDGAN-IM in Appendix. 
\vspace{-0.1cm}
\paragraph{SPGDGAN-IM} To obtain a theoretical insight effectively in the PGD-based framework, we additionally consider a sparsity-promoting operation in iterations of PGDGAN and PGDGAN as follows, respectively. This consideration makes it possible in practice as the image can be viewed in general as a sparse representation under a certain unitary transform domain (e.g., the wavelet transform domain) and there have been rich theoretical backgrounds/guarantees \citep{lee2012subspace,bora2017compressed,dhar2018modeling,jagatap2019algorithmic} for recovering sparse signal in CS.
\begin{align}\label{spgd_test}
    \boldsymbol{z}_{t+1} &= \underset{\boldsymbol{z}}{\arg\min}\left\| \boldsymbol{x}_{t} - \alpha  \boldsymbol{A}^{\top}(\boldsymbol{A} \boldsymbol{x}_{t} - \boldsymbol{y}_{te}) - G_{\boldsymbol{\theta}^*}(\boldsymbol{z})\right\|, \nonumber \\
    \boldsymbol{x}_{t+1} &=\boldsymbol{U}h_{s}\big(\boldsymbol{U}^{\top} G_{\boldsymbol{\theta}^*}(\boldsymbol{z}_{t+1})\big),\\\label{spgdim_test}
    \boldsymbol{z}_{t+1} &= \underset{\boldsymbol{z}}{\arg\min}\left\| \boldsymbol{x}_{t} - \alpha  \boldsymbol{A}^{\top}(\boldsymbol{A} \boldsymbol{x}_{t} - \boldsymbol{y}_{te}) - G_{\boldsymbol{\theta}^*}(\boldsymbol{z},\boldsymbol{y}_{te})\right\|, \nonumber \\
    \boldsymbol{x}_{t+1} &= \boldsymbol{U}h_{s}\big(\boldsymbol{U}^{\top} G_{\boldsymbol{\theta}^*}(\boldsymbol{z}_{t+1},\boldsymbol{y}_{te})\big),
\end{align}
where $\boldsymbol{U}$ is an unitary transform matrix to reveal the sparsity of signals, $h_s(\boldsymbol{v})$ is a hard-thresholding to remain the $s$-largest elements of $\boldsymbol{v}$ otherwise forcing to zero. Thus, \eqref{spgd_test} and \eqref{spgdim_test} are the iterations modified from those of PGDGAN and PGDGAN-IM to make the generated model output a sparse signal in the transform domain, respectively. We dub these modified frameworks sparse PGDGAN and PGDGAN-IM (SPGDGAN and SPGDGAN-IM), respectively. 

\section{Theoretical Insight}

Based on SPGDGAN in Section \ref{subsec:pgd}, we provide a theoretical insight for the reason why inserting measurements into generative models improves the performance of signal reconstruction (i.e., why SPGDGAN-IM outcompetes SPGDGAN). Both methods require $\boldsymbol{A}$ to satisfy $(\mathcal{S},1-\gamma,1+\gamma)$-RIP. 
\begin{defn} For a parameter $\gamma>0$, $\boldsymbol{A} \in \mathbb{R}^{m \times d}$ satisfies $(\mathcal{S},1-\gamma,1+\gamma)$-RIP, if for all $\boldsymbol{x} \in \mathcal{S}$,
\begin{align}\label{rip_cond}
    (1-\gamma)  \left\| \boldsymbol{x}  \right\|  \leq \left\| \boldsymbol{Ax}  \right\| \leq (1+\gamma)  \left\| \boldsymbol{x}  \right\|.
\end{align}
\end{defn}
Under \eqref{rip_cond}, we present a condition for SPGDGAN or SPGDGAN-IM to recover signals as follows.

\begin{thm} \label{thm_im}  Let $\{\boldsymbol{x}_t\}_{t=0}^{T}$  be a set of outputs obtained from each iteration of SPGDGAN or SPGDGAN-IM. 
Let $\boldsymbol{x}_{te}$  and  $\boldsymbol{y}_{te}$ be a target signal vector and its measurement vector, respectively. Let $\boldsymbol{x}_{te}$ be a $s$-sparse vector in an unitary transform domain, i.e., $\boldsymbol{x}_{te}=\boldsymbol{U}h_s(\boldsymbol{U}^{\top}\boldsymbol{x}_{te})$ with an arbitrary unitary matrix $\boldsymbol{U}$. Define set $\Sigma:=\{\Gamma^{(1)}_{t}\}_{t=0}^{T-1} \cup \{\Gamma^{(2)}_{t}\}_{t=0}^{T-1}$ where $\Gamma^{(1)}_{t} := \textup{supp}(h_s(\boldsymbol{U}^{\top}\boldsymbol{x}_t) -  h_s(\boldsymbol{U}^{\top}\boldsymbol{x}_{te}))$ and $\Gamma^{(2)}_{t} := \textup{supp}(h_s(\boldsymbol{U}^{\top}\boldsymbol{x}_{t+1}) -  h_s(\boldsymbol{U}^{\top}\boldsymbol{x}_{t}))$ for $t \in \{0:T-1\}$. Let $\boldsymbol{A}\boldsymbol{U} \in \mathbb{R}^{m \times d}$ satisfy  $(\mathcal{S},1-\gamma,1+\gamma)$-RIP with high probability, where $\mathcal{S}:= \{\boldsymbol{x} | \textup{supp}(\boldsymbol{x}) \in \Sigma \}$. Suppose 
\begin{enumerate}[label=(\alph*)]
    \item In the case for SPGDGAN, there exists a vector $\boldsymbol{z}^*$ satisfying $G_{\boldsymbol{\theta}^*}(\boldsymbol{z}^*)=\boldsymbol{x}_{te}$,\label{condition-a}
    \item In the case for SPGDGAN-IM, there exists a vector $\boldsymbol{z}^*$ satisfying $G_{\boldsymbol{\theta}^*}(\boldsymbol{z}^*,\boldsymbol{y}_{te})=\boldsymbol{x}_{te}$.\label{condition-b}  
\end{enumerate}
Then, there exists $T \propto \log (1/\epsilon)$ such that the signal estimate $ \hat{\boldsymbol{x}} = \boldsymbol{x}_{T}$ satisfies  $\left\| \hat{\boldsymbol{x}} - \boldsymbol{x}_{te} \right\| \leq \epsilon$. 
\end{thm}

\begin{table}[t]
	\caption{Mean and standard deviation of probabilities to satisfy the condition \ref{condition-a} or \ref{condition-b} in Theorem \ref{thm_im} by sampling $1000$ $\boldsymbol{z}$'s for $64$ random test samples. $\boldsymbol{\theta}^*$ in \ref{condition-a} and \ref{condition-b} comes from \eqref{dcgan_obj} and \eqref{im_dcgan_obj}. Here, $m = 1000 \text{ and } \epsilon = 0.125$.} 
	\centering
	{\small
	\resizebox{0.5\linewidth}{!}{
	\begin{tabular}{lc}
		\toprule 
		 $P_{\boldsymbol{z}}\big(\left\lVert G_{\boldsymbol{\theta^*}}(\boldsymbol{z}) - \boldsymbol{x}_{te} \right\rVert^2 < \epsilon \big)$ & $0.003 \pm 0.007$ \\[0.1cm] $P_{\boldsymbol{z}}\big(\left\lVert G_{\boldsymbol{\theta^*}}(\boldsymbol{z}, \boldsymbol{y}_{te}) - \boldsymbol{x}_{te} \right\rVert^2 < \epsilon \big)$ & $\mathbf{0.984 \pm 0.124}$  \\
		\bottomrule
	\end{tabular}
	}
	} 
	\label{table_th}
\end{table}

Theorem \ref{thm_im} shows that the condition required for successful signal recovery is changed from (a) to (b), with the introduction of IM into SPGDGAN. Both conditions (a) and (b) require that the target signal to recover is included in the range of a pre-trained generator, but we empirically show in Table \ref{table_th} that the condition (b) with IM is much easier to be satisfied than (a), which is consistent with our motivation introduced in Section \ref{subsec:framework}. Overall, IM improves the performance of signal recovery by relaxing the condition for signal existence from (a) to (b). 

The question of how easily the RIP condition required in Theorem \ref{thm_im} is satisfied in practice will naturally arise. Theorem \ref{thm_rip} ensures that it holds in the standard CS case with Gaussian $\boldsymbol{A}$ if the number of measurements $m$ follows the order of sparsity $s$ (i.e., $m = \Omega(s)$) as given in \eqref{mbound}. 

\begin{thm}\label{thm_rip}
Given $s,m,d \in \mathbb{N}$ satisfying $2s \leq m \leq d$, consider a random matrix $\boldsymbol{A} \in \mathbb{R}^{m \times d}$ whose entries
are i.i.d. Gaussian following $\mathcal{N}(0,1/m)$.
Then, for $\mathcal{S}$ given in Theorem \ref{thm_im}, $\boldsymbol{AU}$ satisfies $(\mathcal{S},1-\gamma,1+\gamma)$-RIP with probability at least $1-\tau$ for any $\tau$ and $\gamma$, if
\begin{align}\label{mbound} 
&m \geq \frac{2 ({s+  \ln(4T/\tau})}{(\sqrt{1+\gamma}-1)^2}. 
\end{align}
\label{thm2}
\end{thm}
\vspace{-0.5cm}

It is instructive to note that the proof of our theory \ref{thm_im} is based on that of \citet[Theorem 1]{jagatap2019algorithmic}, but is non-trivially modified. In fact, the intermediate states of gradient descent in the proof of \citet[Theorem 1]{jagatap2019algorithmic} should satisfy a RIP-like condition (e.g., REC), but they actually do not. To solve this problem, we use the original RIP and further consider the sparsity-promoting operator (i.e., \eqref{spgd_test} and \eqref{spgdim_test}) for these states to have a finite sparsity (i.e., satisfy RIP). This conversion makes the target matrix that must satisfy RIP become not $\boldsymbol{A}$ but $\boldsymbol{AU}$, where $\boldsymbol{U}$ denotes an unitary transform operator. However, given that $\boldsymbol{AU}$ also has the same statistical characteristics as $\boldsymbol{A}$ (when $\boldsymbol{A}$ follows Gaussian distribution, i.e., standard setup for $\boldsymbol{A}$ in CS), just as $\boldsymbol{A}$ easily satisfies RIP, $\boldsymbol{AU}$ also easily satisfies RIP as we prove it in Theorem \ref{thm2}.

\begin{figure*}[t]
    \vskip -5pt
	\centering
	\subfigure[The first test sample]{\includegraphics[width=0.49\linewidth ]{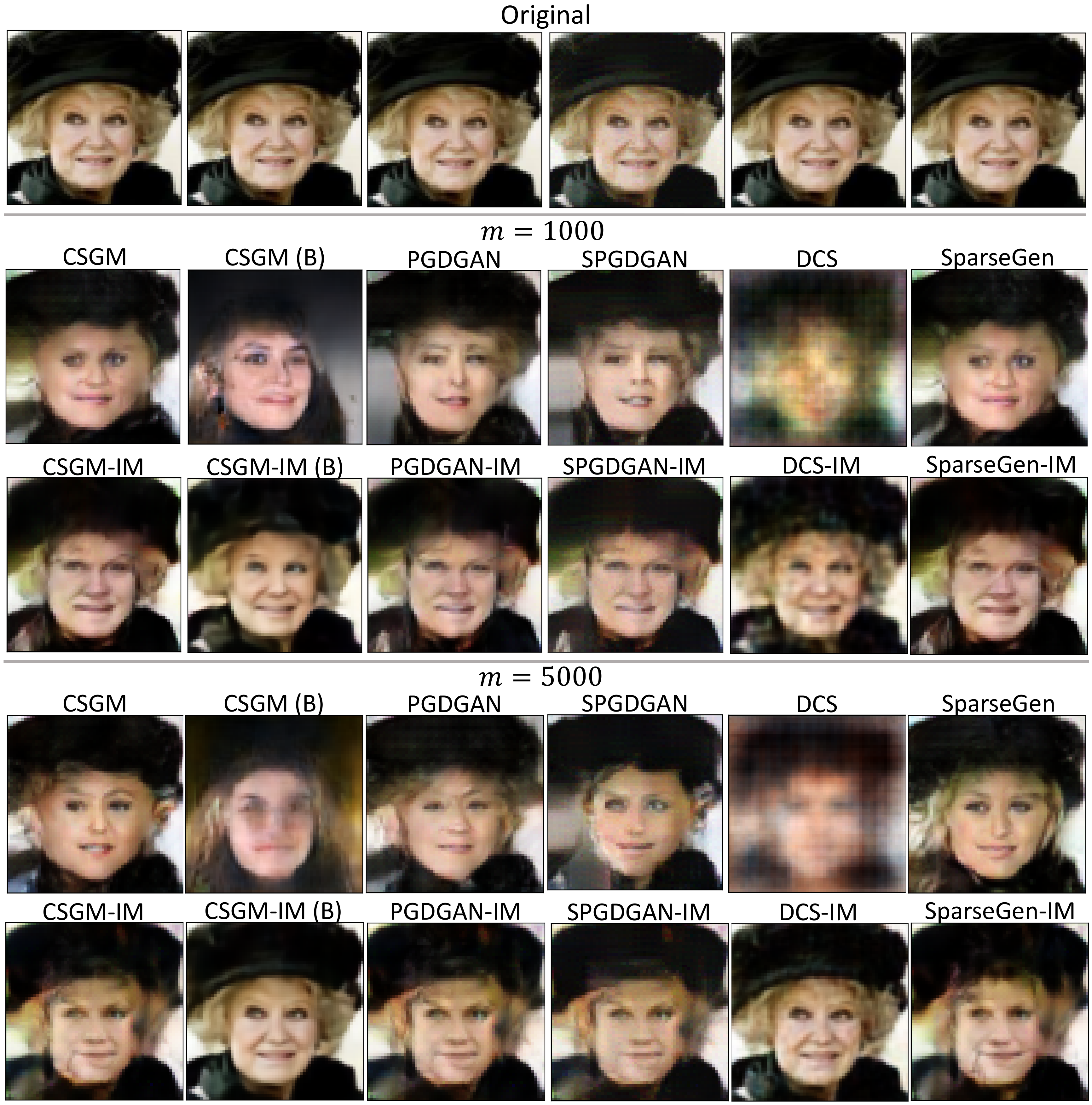}}
	\subfigure[The second test sample]{\includegraphics[width=0.49\linewidth ]{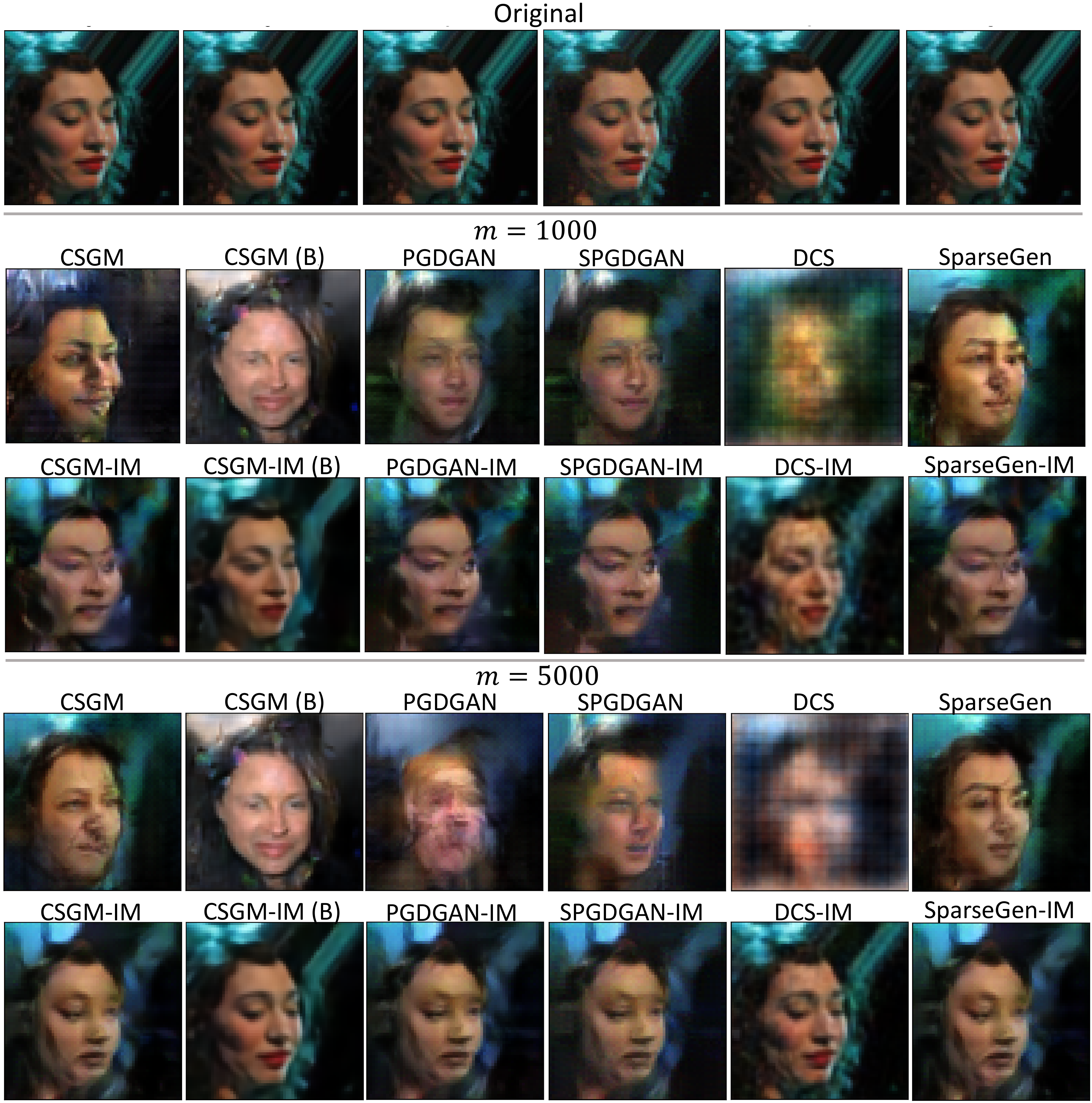}}
	\vskip -8pt
	\caption{\footnotesize Reconstructed images on CelebA when $m = 1000 \,\,(m/d \approx 8.14 \%)$ and $m = 5000 \,\, (m/d \approx 40.69 \%)$. The first row represents the original images. Note that the original image in the fourth column of each test sample is a $s$-sparse vector in the wavelet basis when $s=d/2$. The second and third row indicate images recovered by existing methods and existing methods + IM  when $m = 1000$. The fourth and last rows display images recovered by existing methods and existing methods + IM  when $m = 5000$.}
	\label{fig:celeba}
	\vspace{-0.2cm}
\end{figure*}

\begin{table*}[t]
    \vskip -5pt 
	\caption{\footnotesize Reconstruction error per pixel with $95\%$ confidence interval of $5$ trials for existing methods and those with the IM framework. DCGAN is employed unless ``B'' is marked, where ``B'' stands for the use of BEGAN.}
	\scriptsize
	\centering
	\vskip -5pt 
	{
	\begin{tabular}{lccccc}
		\toprule
		Method & $m=20$ & $m=100$ & $m=500$ & $m=1000$ & $m=5000$ \\
		\midrule
		CSGM \citep{bora2017compressed} & $0.304 \pm 0.068$ & $0.104 \pm 0.012$ & $0.039 \pm 0.004$ & $0.033 \pm 0.003$ & $0.029 \pm 0.003$ \\[0.1cm]
		CSGM-IM & $\mathbf{0.209 \pm 0.011}$ & $\mathbf{0.072 \pm 0.007}$ & $\mathbf{0.029 \pm 0.003}$ & $\mathbf{0.022 \pm 0.003}$ & $\mathbf{0.018 \pm 0.002}$ \\
		\midrule
		CSGM (B) & $0.213 \pm 0.031$ & $0.104 \pm 0.012$ & $0.071 \pm 0.011$ & $0.068 \pm 0.010$ & $0.066 \pm 0.011$ \\[0.1cm]
		\makecell[l]{CSGM-IM (B)} & $\mathbf{0.185 \pm 0.008}$ & $\mathbf{0.058 \pm 0.007}$ & $\mathbf{0.015 \pm 0.002}$ & $\mathbf{0.008 \pm 0.001}$ & $\mathbf{0.002 \pm 0.000}$ \\
		\midrule
		PGDGAN \citep{shah2018solving} & $0.630 \pm 0.076$ & $0.128 \pm 0.010$ & $0.049 \pm 0.006$ & $0.038 \pm 0.004$ & $0.032 \pm 0.004$ \\[0.1cm]
	    PGDGAN-IM & $\mathbf{0.438 \pm 0.017}$ & $\mathbf{0.094 \pm 0.007}$ & $\mathbf{0.031 \pm 0.003}$ & $\mathbf{0.023 \pm 0.003}$ & $\mathbf{0.019 \pm 0.002}$ \\
	    \midrule
		SPGDGAN (Ours) & $0.624 \pm 0.110$ & $0.124 \pm 0.010$ & $0.048 \pm 0.005$ & $0.039 \pm 0.004$ & $0.032 \pm 0.004$ \\[0.1cm]
		SPGDGAN-IM  & $\mathbf{0.425 \pm 0.021}$ & $\mathbf{0.090 \pm 0.008}$ & $\mathbf{0.031 \pm 0.003}$ & $\mathbf{0.024 \pm 0.002}$ & $\mathbf{0.020 \pm 0.002}$ \\
		\midrule
		DCS \citep{wu2019deep} & $0.246 \pm 0.005$ & $0.159 \pm 0.007$ & $0.114 \pm 0.007$ & $0.087 \pm 0.003$ & $0.082 \pm 0.002$ \\[0.1cm]
	    DCS-IM & $\mathbf{0.110 \pm 0.008}$ & $\mathbf{0.049 \pm 0.005}$ & $\mathbf{0.017 \pm 0.002}$ & $\mathbf{0.010 \pm 0.001}$ & $\mathbf{0.002 \pm 0.000}$ \\
	    \midrule
		SparseGen \citep{dhar2018modeling} & $0.374 \pm 0.060$ & $0.118 \pm 0.018$ & $0.038 \pm 0.005$ & $0.030 \pm 0.004$ & $0.024 \pm 0.003$ \\[0.1cm]
	    SparseGen-IM & $\mathbf{0.224 \pm 0.022}$ & $\mathbf{0.072 \pm 0.005}$ & $\mathbf{0.028 \pm 0.003}$ & $\mathbf{0.021 \pm 0.002}$ & $\mathbf{0.017 \pm 0.002}$ \\
		\bottomrule
	\end{tabular}
	}
	\vspace{-0.2cm}
	\label{tab:total_experiments}
\end{table*}

\section{Experiments}
\subsection{Comparison to Existing CSPGs}\label{exp}
In order to evaluate the effectiveness of our algorithm, we focus on conducting experiments on CelebA \citep{liu2015celeba} dataset, which is a common but more difficult task than on MNIST \citep{lecun1998mnist} or OMNIGLOT \citep{lake2015omniglot}. 
The images are cropped at the center to the size $64 \times 64 \times 3 \, (d = 12288)$ and normalized into the range $[-1, 1]$.  For inference, we utilize 64 random images in the test set and compute the reconstruction error $\left\lVert \boldsymbol{x}_{te} - \hat{\boldsymbol{x}} \right\rVert^2$ with $95\%$ confidence interval of $5$ trials. 
Each entry of $\boldsymbol{A}$ is sampled from the normal distribution $\mathcal{N}(0, 1/m)$. 

When adding $\boldsymbol{y}$ into generative models, $\boldsymbol{y}$ is concatenated to $\boldsymbol{z}$ in $G_{\boldsymbol{\theta}}$ in both DCGAN and BEGAN. $\boldsymbol{y}$ is also concatenated to the embedding layer in $D_{\boldsymbol{\phi}}$ in BEGAN. Owing to the absence of an embedding layer in $D_{\boldsymbol{\phi}}$ in DCGAN, we emulate an architecture in \citet{reed2016textgan}. The dimension of $\boldsymbol{z}$ is set to $100$ in DCGAN and $64$ in BEGAN, respectively. 
Further information about experimental settings and more experiments on different $m$ are postponed to Appendix.


\vspace{-0.15cm}
\paragraph{CSGM-IM}
As illustrated in Table \ref{tab:total_experiments}, IM decreases the reconstruction error per pixel of CSGM by above $30\%$ on average 
, but CSGM-IM still exhibits performance saturation like CSGM. When exploiting BEGAN instead of DCGAN, however, CSGM-IM achieves better performance 
as well as overcomes such a limitation. 
Remarkably, for $m \ge 1000$, the reconstruction error per pixel of CSGM-IM using BEGAN is almost less than that of CSGM using BEGAN by an order of magnitude, which makes CSGM-IM using BEGAN reconstruct images similar to the original ones as shown in Figure \ref{fig:celeba}. 


\vspace{-0.15cm}
\paragraph{PGDGAN-IM} 
For fair comparisons, we use the same hyperparameters as \citet{shah2018solving}. In Table \ref{tab:total_experiments}, IM reduces the reconstruction error per pixel of PGDGAN by above $30\%$ overall. 

\vspace{-0.15cm}
\paragraph{SPGDGAN-IM}
Following \citet{shah2018solving}, SPGDGAN and SPGDGAN-IM with $s=d/2$ are on par with PGDGAN and PGDGAN-IM respectively despite the presence of a sparsity operator.


\vspace{-0.15cm}
\paragraph{DCS-IM} 
To verify the validness of our method in the absence of $D_{\boldsymbol{\phi}}$, we apply IM to DCS. 
Table \ref{tab:total_experiments} shows that IM lowers the reconstruction error per pixel of DCS by above $70\%$ on average and by an order of magnitude for $m \ge 1000$, which leads to successful signal recovery like CSGM-IM using BEGAN in Figure \ref{fig:celeba}. More noticeably, DCS-IM outperforms all the other methods when $m \le 100$. 

\vspace{-0.15cm}
\paragraph{SparseGen-IM}
We only consider the wavelet basis due to the fact that \citet{dhar2018modeling} recommend the wavelet basis rather than the discrete cosine transform. Table \ref{tab:total_experiments} indicates that SparseGen-IM surpasses SparseGen, averagely curtailing the reconstruction error by around $38\%$. 

\subsection{Ablation Studies}\label{ablation} 

\begin{figure*}
\centering
	\subfigure[$m=100$\label{figiter_a}]{\includegraphics[width=0.24\linewidth, height = 3cm]{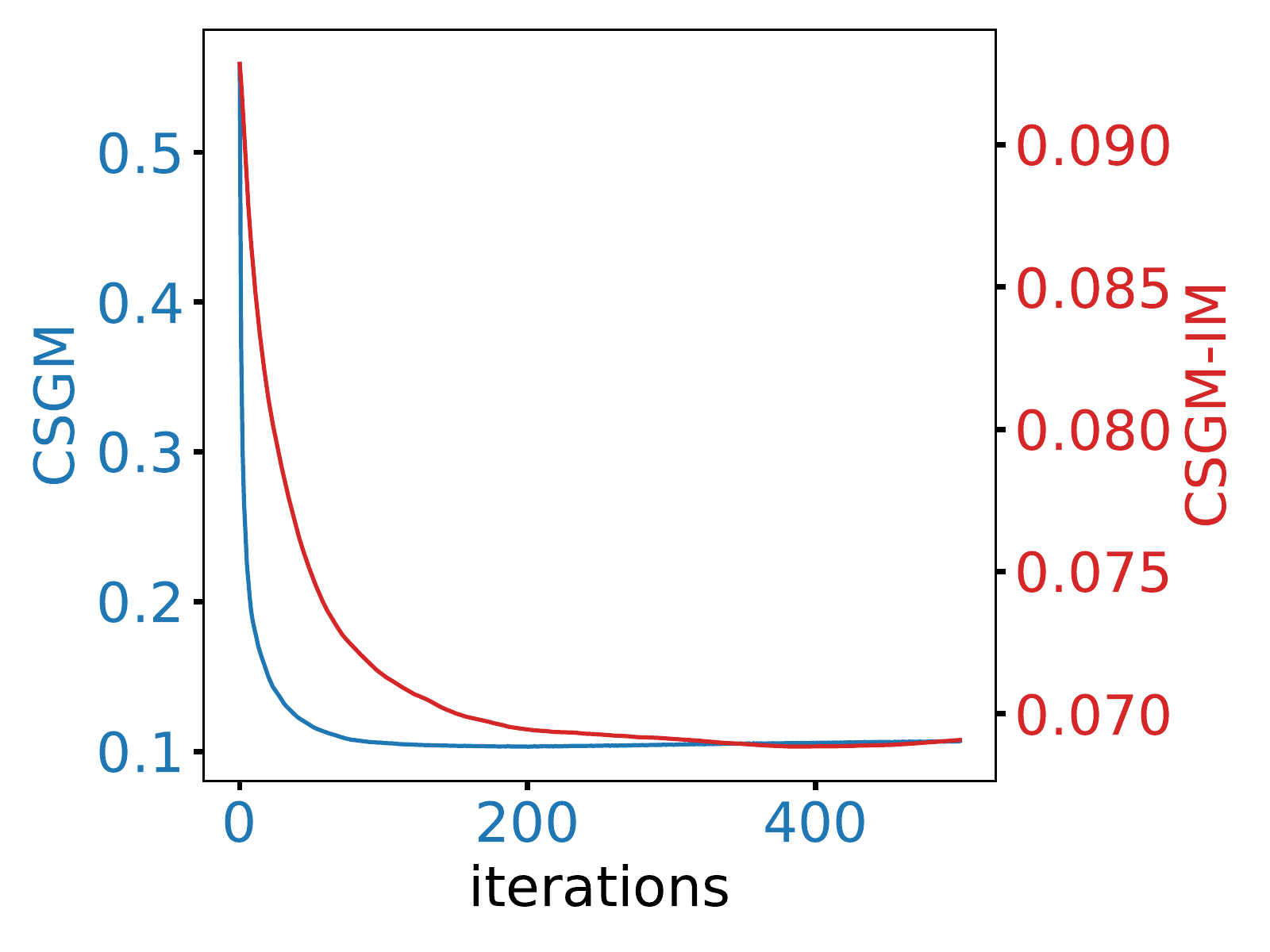}}
	\subfigure[$m=2500$\label{figiter_b}]{\includegraphics[width=0.24\linewidth, height = 3cm]{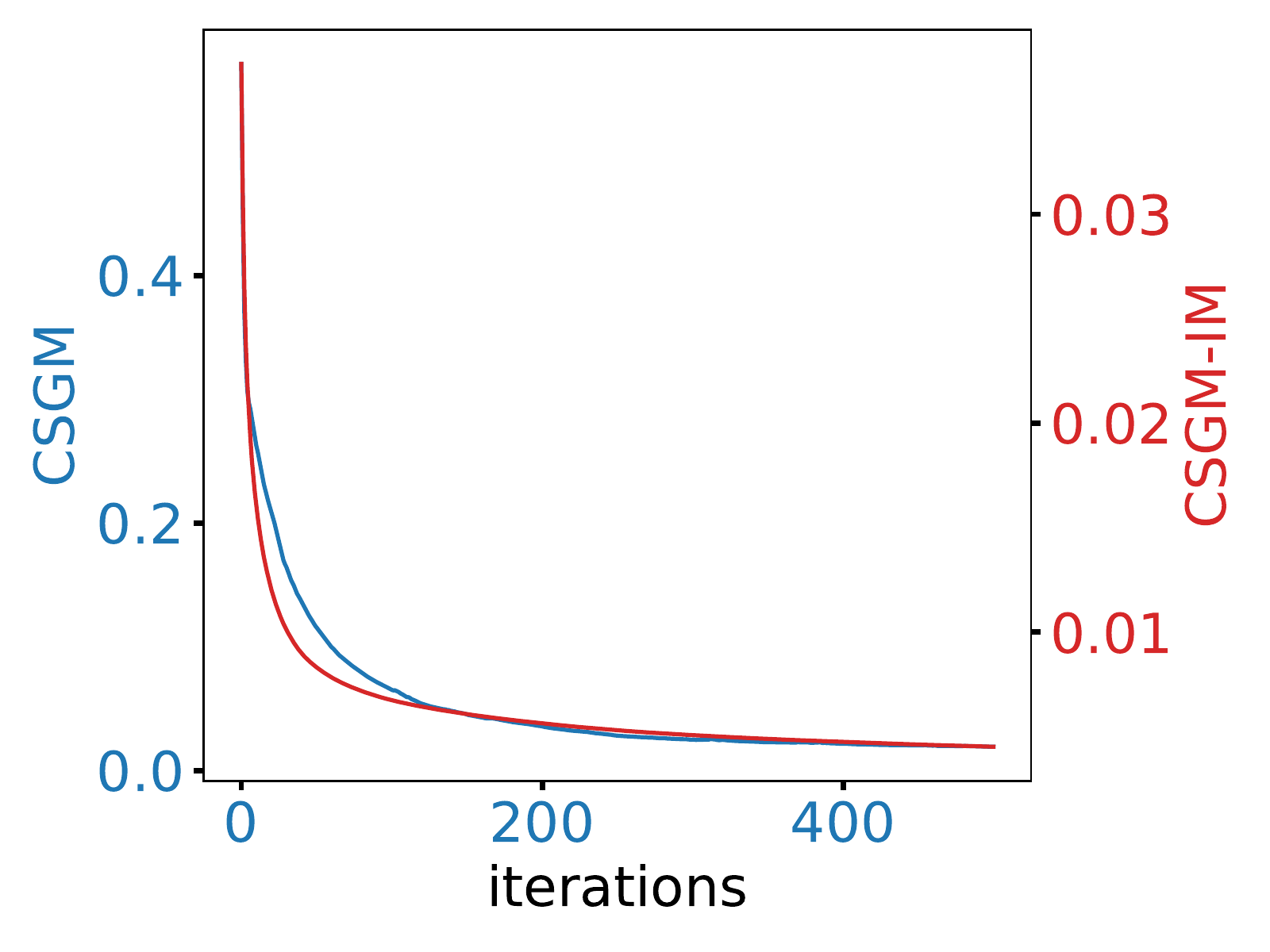}}
	\subfigure[$d_{\boldsymbol{z}_{\text{CSGM-IM}}}=100$\label{figiter_c}]{\includegraphics[width=0.24\linewidth, height = 3cm]{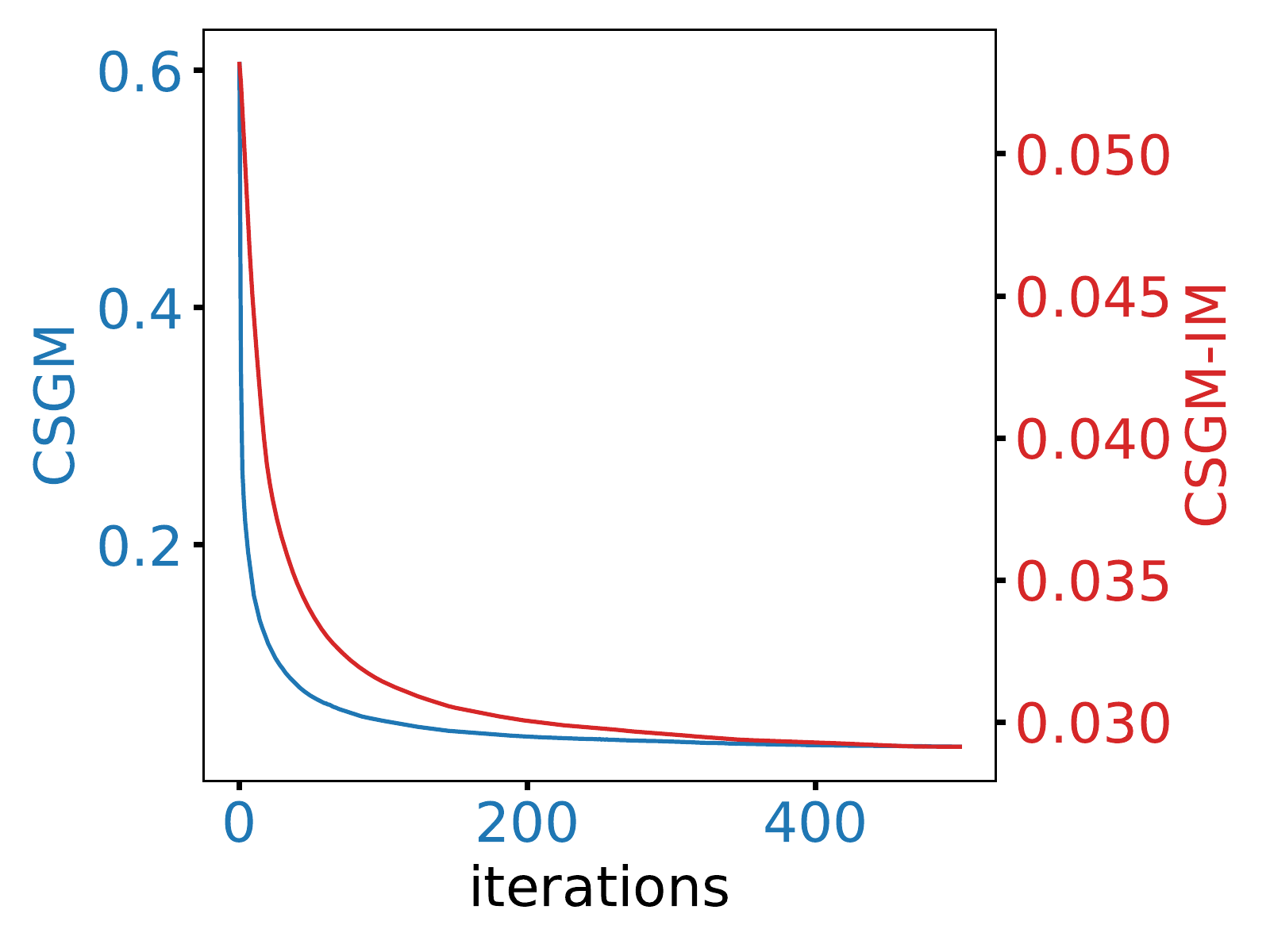}}
	\subfigure[$d_{\boldsymbol{z}_{\text{CSGM-IM}}}=2500$\label{figiter_d}]{\includegraphics[width=0.24\linewidth, height = 3cm]{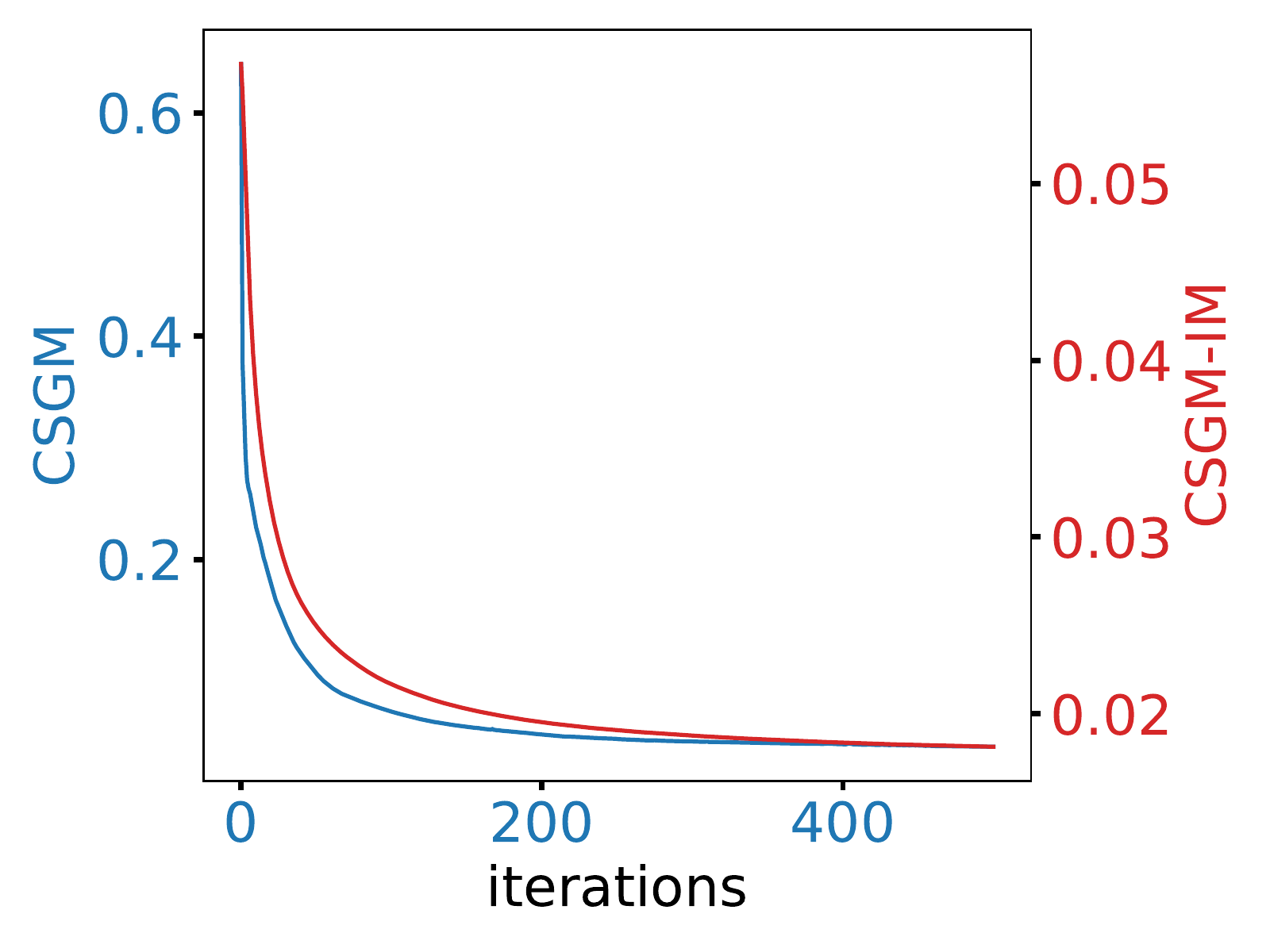}}
	\caption{\footnotesize Test phases of CSGM (the blue curve) and CSGM-IM (the red curve) -- (a) and (b) for results in Table \ref{tab:same_weights}, (c) and (d) for results in  Table \ref{tab:m_fixed}.}
	\label{fig:same_weights}
	\vspace{-0.2cm}
\end{figure*}

To figure out whether the improvement of IM merely results from the increased number of weights in the generator of CSGM-IM or not, we examine the case when the dimension of $\boldsymbol{z}$ in CSGM equals the sum of $m$ and that in CSGM-IM so as to make the number of weights in the generator of CSGM the same as that in the generator of CSGM-IM. In other words, $d_{\boldsymbol{z}_{\text{CSGM}}} = m + d_{\boldsymbol{z}_{\text{CSGM-IM}}}$, where $d_{\boldsymbol{z}_{\text{CSGM}}}$ and $d_{\boldsymbol{z}_{\text{CSGM-IM}}}$ are the dimension of $\boldsymbol{z}_{\text{CSGM}}$, $\boldsymbol{z}$ in CSGM, and that of $\boldsymbol{z}_{\text{CSGM-IM}}$, $\boldsymbol{z}$ in CSGM-IM. Here, we let $d_{\boldsymbol{z}_{\text{CSGM-IM}}}$ equal $m$ to balance them. Table \ref{tab:same_weights} shows there is still performance gap for every $m$ even if the generator of CSGM has the same number of weights as that of CSGM-IM. Moreover, the latent optimization performs well in CSGM-IM as seen in Figures \ref{figiter_a} and \ref{figiter_b}.

\begin{table}[t]
    \parbox{.49\linewidth}{
	\caption{\footnotesize {Reconstruction error per pixel with $95\%$ confidence interval of $5$ trials for Section \ref{subsec:csgm}, where
    the number of weights in the generator of CSGM equals that in the generator of CSGM-IM and $d_{\boldsymbol{z}_{\text{CSGM-IM}}}$ equals $m$.}} 
	\centering
	{
	\begin{tabular}{lcc}
		\toprule
		$m$ & CSGM & CSGM-IM \\
		\midrule
		$100$ & $0.103 \pm 0.014$ & $\mathbf{0.072 \pm 0.007}$ \\
		$500$ & $0.029 \pm 0.003$ & $\mathbf{0.019 \pm 0.002}$ \\
		$2500$ & $0.018 \pm 0.002$ & $\mathbf{0.005 \pm 0.000}$ \\
		\bottomrule
	\end{tabular}
	}
	\label{tab:same_weights} 
    } 
    \parbox{.49\linewidth}{
	\caption{\footnotesize {Reconstruction error per pixel with $95\%$ confidence interval of $5$ trials for Section \ref{subsec:csgm}, where the number of weights in generator of CSGM equals that in generator of CSGM-IM with $m=500$ and $d_{\boldsymbol{z}_{\text{CSGM-IM}}}$ varies.}}
	\centering
	{
	\begin{tabular}{lcc}
		\toprule
		$d_{\boldsymbol{z}_{\text{CSGM-IM}}}$ & CSGM & CSGM-IM \\
		\midrule
		$100$ & $0.030 \pm 0.004$ & $\mathbf{0.028 \pm 0.003}$ \\
		$500$ & $0.029 \pm 0.003$ & $\mathbf{0.019 \pm 0.003}$ \\
		$2500$ & $0.031 \pm 0.007$ & $\mathbf{0.019 \pm 0.003}$ \\
		\bottomrule
	\end{tabular}
	}
	\label{tab:m_fixed}
    }
\end{table}

To verify the superiority of CSGM-IM to CSGM when the ratio of $m$ to  $d_{\boldsymbol{z}_{\text{CSGM-IM}}}$ varies but still $d_{\boldsymbol{z}_{\text{CSGM}}} = m + d_{\boldsymbol{z}_{\text{CSGM-IM}}}$, we conduct experiments on different $d_{\boldsymbol{z}_{\text{CSGM-IM}}}$ with $m$ fixed to $500$. 
As seen in Table \ref{tab:m_fixed} and Figures \ref{figiter_c} and \ref{figiter_d}, CSGM-IM surpasses CSGM as well as the latent optimization works well in CSGM-IM no matter how small or large $d_{\boldsymbol{z}_{\text{CSGM-IM}}}$ is. Hence, the direct insertion of $\boldsymbol{y}$ into generators plays a vital role in performance enhancement.  

\section{Application to Magnetic Resonance Imaging}
In this section, we run experiments on knee data in fastMRI \citep{zbontar2018fastmri} to validate the practicality of IM on real-world data. 
Owing to the low quality of test slices, 
we utilize $64$ random images in the validation slices for inference. 
Unlike \eqref{CS}, noise is not added when creating a measurement vector. In this experiment, we solely employ BEGAN due to the fact that it works best among our experiments.
The details of the experimental setup are deferred to Appendix.

\begin{figure}[t]
    \vskip -3pt
	\centering
	\subfigure[The $193$rd validation slice]{\includegraphics[width=0.49\linewidth]{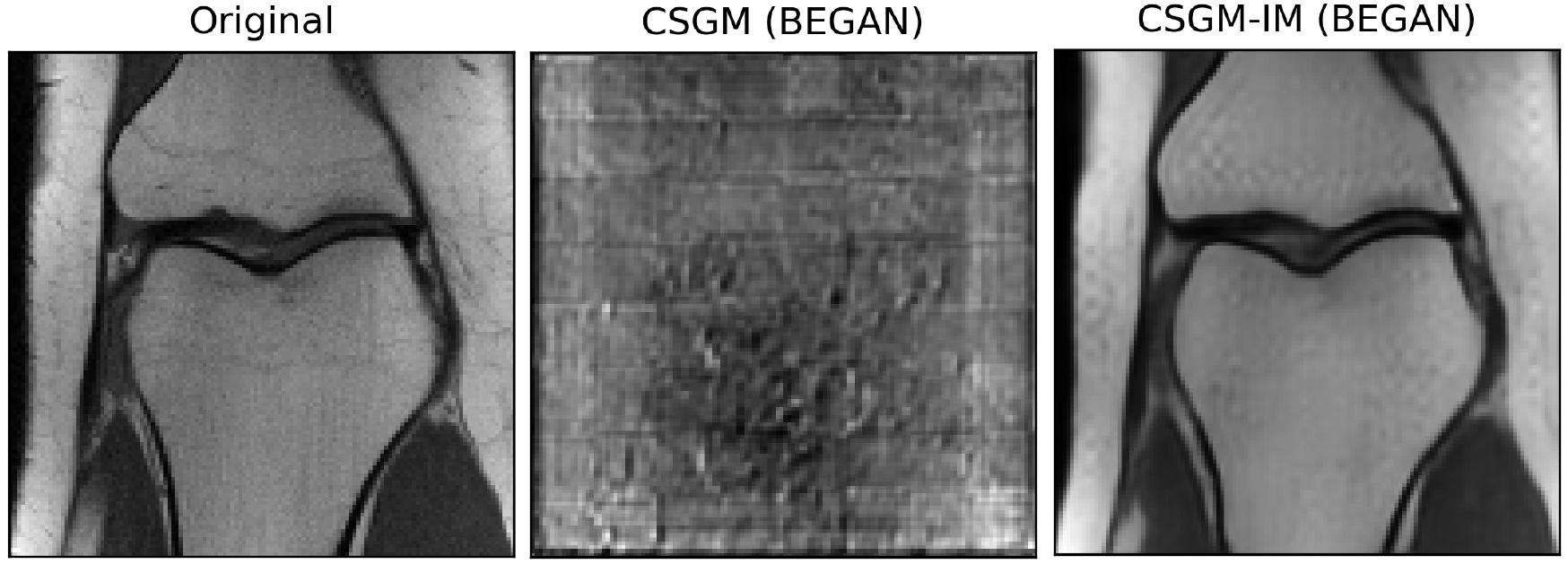}}
	\subfigure[The $928$th validation slice]{\includegraphics[width=0.49\linewidth]{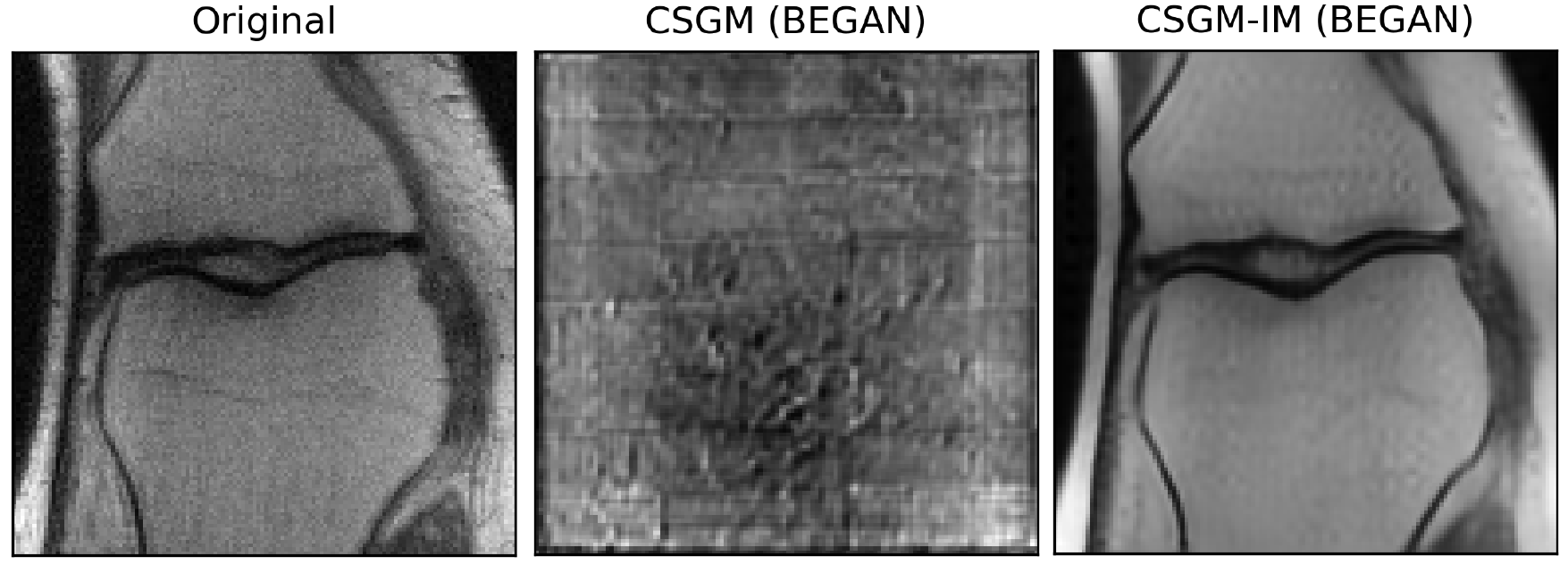}}
	\vskip -8pt
	\caption{\footnotesize Reconstructed images on fastMRI when $m = 5000 \, (m/d \approx 30.52\%)$. The first column has the original image. The second and last column show images recovered by CSGM and CSGM-IM using BEGAN.} 
	\label{fig:mri_m5000}
	\vspace{-0.2cm}
\end{figure}

Similarly to Section \ref{exp}, CSGM-IM outdoes 
CSGM as seen in Table \ref{tab:MRI_BEGAN}. For $m \ge 2500$, IM lowers the reconstruction error of CSGM by closely an order of magnitude. Not only that, as shown in Figure \ref{fig:mri_m5000}, CSGM-IM recovers images highly analogous to the original ones whereas CSGM cannot at all.


\section{Conclusion}

\begin{table*} 
	\caption{\footnotesize MRI reconstruction error per pixel with $95\%$ confidence interval of $5$ trials for Section \ref{subsec:csgm} with BEGAN.}
	\vskip -5pt
	\centering
	{
	\begin{tabular}{lcc}
		\toprule
		$m$ & CSGM (B) & CSGM-IM (B) \\
		\midrule
		$500$ & $0.0674 \pm 0.0118$ & $\mathbf{0.0159 \pm 0.0007}$ \\
		$1000$ & $0.0671 \pm 0.0097$ & $\mathbf{0.0120 \pm 0.0007}$ \\
		$2500$ & $0.0655 \pm 0.0101$ & $\mathbf{0.0092 \pm 0.0005}$ \\
		$5000$ & $0.0659 \pm 0.0123$ & $\mathbf{0.0082 \pm 0.0005}$ \\
		\bottomrule
	\end{tabular}
	}
	\label{tab:MRI_BEGAN}
	\vspace{-0.2cm}
\end{table*}

We propose a simple yet effective method, \emph{Inserting Measurements}, which allows a generator to exploit $\boldsymbol{y}$ in both of training and test phases, where existing generative models are limited to use $\boldsymbol{y}$ only in the test phase. Even in the IM framework, the characteristic of generative models remains, which allows us to find a more closer estimate to the true signal by the latent optimization. By leveraging both advantages of discriminative and generative models, IM can yield much smaller reconstruction error than existing methods up to an order of magnitude. We therefore expect IM to be useful for a variety of CS applications as well as image recovery.

{\small
\bibliographystyle{unsrtnat}
\bibliography{reference}

\begin{thebibliography}{46}
\providecommand{\natexlab}[1]{#1}
\providecommand{\url}[1]{\texttt{#1}}
\expandafter\ifx\csname urlstyle\endcsname\relax
  \providecommand{\doi}[1]{doi: #1}\else
  \providecommand{\doi}{doi: \begingroup \urlstyle{rm}\Url}\fi

\bibitem[He et~al.(2018)He, Wen, Jin, and Li]{he2018deep}
Hengtao He, Chao-Kai Wen, Shi Jin, and Geoffrey~Ye Li.
\newblock Deep learning-based channel estimation for beamspace mm{W}ave massive
  {MIMO} systems.
\newblock \emph{IEEE Wireless Communications Letters}, 7\penalty0 (5):\penalty0
  852--855, 2018.

\bibitem[Kim and Chung(2020)]{kim2019tree}
Kyung-Su Kim and Sae-Young Chung.
\newblock Tree search network for sparse estimation.
\newblock \emph{Digital Signal Processing}, 100:\penalty0 102680, 02 2020.
\newblock \doi{10.1016/j.dsp.2020.102680}.

\bibitem[Lustig et~al.(2007)Lustig, Donoho, Santos, and Pauly]{lustig07csmri}
Michael Lustig, David~L. Donoho, Juan~M. Santos, and John~M. Pauly.
\newblock Compressed sensing {MRI}.
\newblock In \emph{IEEE Signal Processing Magazine}, 2007.

\bibitem[Sun et~al.(2016)Sun, Li, Xu, et~al.]{sun2016deep}
Jian Sun, Huibin Li, Zongben Xu, et~al.
\newblock Deep {ADMM}-{N}et for compressive sensing {MRI}.
\newblock In \emph{Advances in Neural Information Processing Systems}, pages
  10--18, 2016.

\bibitem[Willett et~al.(2011)Willett, Marcia, and
  Nichols]{willett2011compressed}
Rebecca~M Willett, Roummel~F Marcia, and Jonathan~M Nichols.
\newblock Compressed sensing for practical optical imaging systems: a tutorial.
\newblock \emph{Optical Engineering}, 50\penalty0 (7):\penalty0 072601, 2011.

\bibitem[Bora et~al.(2017)Bora, Jalal, Price, and Dimakis]{bora2017compressed}
Ashish Bora, Ajil Jalal, Eric Price, and Alexandros~G Dimakis.
\newblock {C}ompressed sensing using generative models.
\newblock In \emph{International Conference on Machine Learning}, pages
  537--546. JMLR. org, 2017.

\bibitem[He et~al.(2017)He, Xin, Ikehata, and Wipf]{he2017bayesian}
Hao He, Bo~Xin, Satoshi Ikehata, and David Wipf.
\newblock From {B}ayesian sparsity to gated recurrent nets.
\newblock In \emph{Advances in Neural Information Processing Systems}, pages
  5554--5564, 2017.

\bibitem[Metzler et~al.(2017)Metzler, Mousavi, and
  Baraniuk]{metzler2017learned}
Chris Metzler, Ali Mousavi, and Richard Baraniuk.
\newblock Learned {D-AMP}: {P}rincipled neural network based compressive image
  recovery.
\newblock In \emph{Advances in Neural Information Processing Systems}, pages
  1772--1783, 2017.

\bibitem[Van~Veen et~al.(2018)Van~Veen, Jalal, Soltanolkotabi, Price,
  Vishwanath, and Dimakis]{van2018compressed}
Dave Van~Veen, Ajil Jalal, Mahdi Soltanolkotabi, Eric Price, Sriram Vishwanath,
  and Alexandros~G Dimakis.
\newblock Compressed sensing with deep image prior and learned regularization.
\newblock \emph{arXiv preprint:1806.06438}, 2018.

\bibitem[Dhar et~al.(2018)Dhar, Grover, and Ermon]{dhar2018modeling}
Manik Dhar, Aditya Grover, and Stefano Ermon.
\newblock {M}odeling sparse deviations for compressed sensing using generative
  models.
\newblock In \emph{International Conference on Machine Learning}, pages
  1222--1231, 2018.

\bibitem[Mardani et~al.(2018)Mardani, Sun, Donoho, Papyan, Monajemi,
  Vasanawala, and Pauly]{mardani2018neural}
Morteza Mardani, Qingyun Sun, David Donoho, Vardan Papyan, Hatef Monajemi,
  Shreyas Vasanawala, and John Pauly.
\newblock Neural proximal gradient descent for compressive imaging.
\newblock In \emph{Advances in Neural Information Processing Systems}, pages
  9573--9583, 2018.

\bibitem[Lunz et~al.(2018)Lunz, {\"O}ktem, and
  Sch{\"o}nlieb]{lunz2018adversarial}
Sebastian Lunz, Ozan {\"O}ktem, and Carola-Bibiane Sch{\"o}nlieb.
\newblock Adversarial regularizers in inverse problems.
\newblock In \emph{Advances in Neural Information Processing Systems}, pages
  8507--8516, 2018.

\bibitem[Mousavi et~al.(2019)Mousavi, Dasarathy, and Baraniuk]{mousavi2018data}
Ali Mousavi, Gautam Dasarathy, and Richard~G Baraniuk.
\newblock A data-driven and distributed approach to sparse signal
  representation and recovery.
\newblock In \emph{International Conference on Learning Representations}, 2019.

\bibitem[Grover and Ermon(2019)]{grover2019uncertainty}
Aditya Grover and Stefano Ermon.
\newblock {U}ncertainty autoencoders: {L}earning compressed representations via
  variational information maximization.
\newblock In \emph{International Conference on Artificial Intelligence and
  Statistics}, pages 2514--2524, 2019.

\bibitem[Wu et~al.(2019{\natexlab{a}})Wu, Dimakis, Sanghavi, Yu, Holtmann-Rice,
  Storcheus, Rostamizadeh, and Kumar]{wu2019learning}
Shanshan Wu, Alexandros Dimakis, Sujay Sanghavi, Felix Yu, Daniel
  Holtmann-Rice, Dmitry Storcheus, Afshin Rostamizadeh, and Sanjiv Kumar.
\newblock Learning a compressed sensing measurement matrix via gradient
  unrolling.
\newblock In \emph{International Conference on Machine Learning}, volume~97,
  2019{\natexlab{a}}.

\bibitem[Wu et~al.(2019{\natexlab{b}})Wu, Rosca, and Lillicrap]{wu2019deep}
Yan Wu, Mihaela Rosca, and Timothy Lillicrap.
\newblock Deep compressed sensing.
\newblock In \emph{International Conference on Machine Learning}, pages
  6850--6860, 2019{\natexlab{b}}.

\bibitem[Raj et~al.(2019)Raj, Li, and Bresler]{raj2019gan}
Ankit Raj, Yuqi Li, and Yoram Bresler.
\newblock {GAN}-based projector for faster recovery with convergence guarantees
  in linear inverse problems.
\newblock In \emph{IEEE/CVF International Conference on Computer Vision}, pages
  5601--5610, 2019.

\bibitem[Tibshirani(1996)]{tibshirani1996lasso}
Robert Tibshirani.
\newblock Regression shrinkage and selection via the {LASSO}.
\newblock \emph{Journal of the Royal Statistical Society: Series B
  (Methodological)}, 58\penalty0 (1):\penalty0 267--288, 1996.

\bibitem[Shah and Hegde(2018)]{shah2018solving}
Viraj Shah and Chinmay Hegde.
\newblock Solving linear inverse problems using {GAN} priors: {A}n algorithm
  with provable guarantees.
\newblock In \emph{IEEE International Conference on Acoustics, Speech and
  Signal Processing}, pages 4609--4613. IEEE, 2018.

\bibitem[Kabkab et~al.(2018)Kabkab, Samangouei, and Chellappa]{kabkab2018task}
Maya Kabkab, Pouya Samangouei, and Rama Chellappa.
\newblock Task-aware compressed sensing with generative adversarial networks.
\newblock In \emph{AAAI Conference on Artificial Intelligence}, 2018.

\bibitem[Gregor and LeCun(2010)]{gregor2010learning}
Karol Gregor and Yann LeCun.
\newblock Learning fast approximations of sparse coding.
\newblock In \emph{International Conference on Machine Learning}, pages
  399--406, 2010.

\bibitem[Moreau and Bruna(2017)]{moreau2017understanding}
Thomas Moreau and Joan Bruna.
\newblock Understanding neural sparse coding with matrix factorization.
\newblock In \emph{International Conference on Learning Representations}, 2017.

\bibitem[Giryes et~al.(2018)Giryes, Eldar, Bronstein, and
  Sapiro]{giryes2018tradeoffs}
Raja Giryes, Yonina~C Eldar, Alex~M Bronstein, and Guillermo Sapiro.
\newblock Tradeoffs between convergence speed and reconstruction accuracy in
  inverse problems.
\newblock \emph{IEEE Transactions on Signal Processing}, 66\penalty0
  (7):\penalty0 1676--1690, 2018.

\bibitem[Chen et~al.(2018)Chen, Liu, Wang, and Yin]{chen2018theoretical}
Xiaohan Chen, Jialin Liu, Zhangyang Wang, and Wotao Yin.
\newblock Theoretical linear convergence of unfolded {ISTA} and its practical
  weights and thresholds.
\newblock In \emph{Advances in Neural Information Processing Systems}, pages
  9061--9071, 2018.

\bibitem[Tramel et~al.(2016)Tramel, Dr{\'e}meau, and
  Krzakala]{tramel2016approximate}
Eric~W Tramel, Ang{\'e}lique Dr{\'e}meau, and Florent Krzakala.
\newblock {A}pproximate message passing with restricted {B}oltzmann machine
  priors.
\newblock \emph{Journal of Statistical Mechanics: Theory and Experiment},
  2016\penalty0 (7):\penalty0 073401, 2016.

\bibitem[Borgerding et~al.(2017)Borgerding, Schniter, and
  Rangan]{borgerding2017amp}
Mark Borgerding, Philip Schniter, and Sundeep Rangan.
\newblock {AMP}-inspired deep networks for sparse linear inverse problems.
\newblock \emph{IEEE Transactions on Signal Processing}, 65\penalty0
  (16):\penalty0 4293--4308, 2017.

\bibitem[Jagatap and Hegde(2019)]{jagatap2019algorithmic}
G~Jagatap and C~Hegde.
\newblock Algorithmic guarantees for inverse imaging with untrained network
  priors.
\newblock In \emph{Advances in Neural Information Processing Systems}, 2019.

\bibitem[Ulyanov et~al.(2018)Ulyanov, Vedaldi, and Lempitsky]{ulyanov2018dip}
Dmitry Ulyanov, Andrea Vedaldi, and Victor Lempitsky.
\newblock Deep image prior.
\newblock In \emph{IEEE/CVF Conference on Computer Vision and Pattern
  Recognition}, pages 9446--9454, 2018.

\bibitem[Heckel et~al.(2019)]{heckel2018deep}
R~Heckel et~al.
\newblock Deep decoder: {C}oncise image representations from untrained
  non-convolutional networks.
\newblock In \emph{International Conference on Learning Representations}, 2019.

\bibitem[Mirza and Osindero(2014)]{mirza2014conditional}
Mehdi Mirza and Simon Osindero.
\newblock Conditional generative adversarial nets.
\newblock \emph{arXiv preprint:1411.1784}, 2014.

\bibitem[Reed et~al.(2016)Reed, Akata, Yan, Logeswaran, Schiele, and
  Lee]{reed2016textgan}
Scott Reed, Zeynep Akata, Xinchen Yan, Lajanugen Logeswaran, Bernt Schiele, and
  Honglak Lee.
\newblock Generative adversarial text to image synthesis.
\newblock In \emph{International Conference on Machine Learning}, pages
  1060--1069, 2016.

\bibitem[Isola et~al.(2017)Isola, Zhu, Zhou, and Efros]{isola2017image}
Phillip Isola, Jun-Yan Zhu, Tinghui Zhou, and Alexei~A Efros.
\newblock Image-to-image translation with conditional adversarial networks.
\newblock In \emph{Proceedings of the IEEE Conference on Computer Vision and
  Pattern Recognition}, pages 1125--1134, 2017.

\bibitem[Wang et~al.(2018)Wang, Liu, Zhu, Tao, Kautz, and
  Catanzaro]{wang2018high}
Ting-Chun Wang, Ming-Yu Liu, Jun-Yan Zhu, Andrew Tao, Jan Kautz, and Bryan
  Catanzaro.
\newblock High-resolution image synthesis and semantic manipulation with
  conditional {GAN}s.
\newblock In \emph{Proceedings of the IEEE Conference on Computer Vision and
  Pattern Recognition}, pages 8798--8807, 2018.

\bibitem[Uelwer et~al.(2019)Uelwer, Oberstra{\ss}, and
  Harmeling]{uelwer2019prcgan}
Tobias Uelwer, Alexander Oberstra{\ss}, and Stefan Harmeling.
\newblock Phase retrieval using conditional generative adversarial networks.
\newblock \emph{arXiv preprint:1912.04981}, 2019.

\bibitem[Ye et~al.(2020)Ye, Liang, Li, and Juang]{ye2020deep}
Hao Ye, Le~Liang, Geoffrey~Ye Li, and Biing-Hwang Juang.
\newblock Deep learning based end-to-end wireless communication systems with
  conditional {GAN} as unknown channel.
\newblock \emph{IEEE Transactions on Wireless Communications}, 2020.

\bibitem[Goodfellow et~al.(2014)Goodfellow, Pouget-Abadie, Mirza, Xu,
  Warde-Farley, Ozair, Courville, and Bengio]{goodfellow2014gan}
Ian Goodfellow, Jean Pouget-Abadie, Mehdi Mirza, Bing Xu, David Warde-Farley,
  Sherjil Ozair, Aaron Courville, and Yoshua Bengio.
\newblock Generative adversarial nets.
\newblock In \emph{Advances in Neural Information Processing Systems}, pages
  2672--2680, 2014.

\bibitem[Radford et~al.(2015)Radford, Metz, and Chintala]{radford2015dcgan}
Alec Radford, Luke Metz, and Soumith Chintala.
\newblock Unsupervised representation learning with deep convolutional
  generative adversarial networks.
\newblock \emph{arXiv preprint:1511.06434}, 2015.

\bibitem[Kingma and Welling(2014)]{kingma2014vae}
Diederik~P. Kingma and Max Welling.
\newblock Auto-encoding variational {B}ayes.
\newblock In \emph{International Conference on Learning Representations}, 2014.

\bibitem[Maaten and Hinton(2008)]{vandermaaten2008tsne}
Laurens van~der Maaten and Geoffrey Hinton.
\newblock Visualizing data using t-sne.
\newblock \emph{Journal of machine learning research}, 9\penalty0
  (Nov):\penalty0 2579--2605, 2008.

\bibitem[Lee et~al.(2012)Lee, Bresler, and Junge]{lee2012subspace}
Kiryung Lee, Yoram Bresler, and Marius Junge.
\newblock Subspace methods for joint sparse recovery.
\newblock \emph{IEEE Transactions on Information Theory}, 58\penalty0
  (6):\penalty0 3613--3641, 2012.

\bibitem[Liu et~al.(2015)Liu, Luo, Wang, and Tang]{liu2015celeba}
Ziwei Liu, Ping Luo, Xiaogang Wang, and Xiaoou Tang.
\newblock Deep learning face attributes in the wild.
\newblock In \emph{Proceedings of International Conference on Computer Vision},
  December 2015.

\bibitem[LeCun et~al.(1998)LeCun, Bottou, Bengio, and Haffner]{lecun1998mnist}
Yann LeCun, L{\'e}on Bottou, Yoshua Bengio, and Patrick Haffner.
\newblock Gradient-based learning applied to document recognition.
\newblock \emph{Proceedings of the IEEE}, 86\penalty0 (11):\penalty0
  2278--2324, 1998.

\bibitem[Lake et~al.(2015)Lake, Salakhutdinov, and Tenenbaum]{lake2015omniglot}
{Brenden M.} Lake, Ruslan Salakhutdinov, and {Joshua B.} Tenenbaum.
\newblock Human-level concept learning through probabilistic program induction.
\newblock \emph{Science}, 350:\penalty0 1332--1338, 2015.

\bibitem[Zbontar et~al.(2018)Zbontar, Knoll, Sriram, Muckley, Bruno, Defazio,
  Parente, Geras, Katsnelson, Chandarana, et~al.]{zbontar2018fastmri}
Jure Zbontar, Florian Knoll, Anuroop Sriram, Matthew~J Muckley, Mary Bruno,
  Aaron Defazio, Marc Parente, Krzysztof~J Geras, Joe Katsnelson, Hersh
  Chandarana, et~al.
\newblock {F}ast{MRI}: {A}n open dataset and benchmarks for accelerated {MRI}.
\newblock \emph{arXiv preprint:1811.08839}, 2018.

\bibitem[Finn et~al.(2017)Finn, Abbeel, and Levine]{finn2017maml}
Chelsea Finn, Pieter Abbeel, and Sergey Levine.
\newblock Model-agnostic meta-learning for fast adaptation of deep networks.
\newblock In \emph{International Conference on Machine Learning}, volume~70,
  pages 1126--1135. PMLR, 2017.

\bibitem[Davidson and Szarek(2001)]{davidson2001local}
Kenneth~R Davidson and Stanislaw~J Szarek.
\newblock {L}ocal operator theory, random matrices and banach spaces.
\newblock \emph{Handbook of the geometry of Banach spaces}, 1\penalty0
  (317-366):\penalty0 131, 2001.

\end{thebibliography}
}

\clearpage
\appendix

\section{Additional Implementations}
\subsection{Deep Compressed Sensing (DCS)}\label{subsec:dsc} 
\label{subsec:dcs}
\paragraph{DCS} To recover signals faster and more accurately than CSGM, \citet{wu2019deep} proposed Deep Compressed Sensing (DCS) by jointly training the latent variables $\boldsymbol{z}$ and the weights $\boldsymbol{\theta}$ of a generator without any discriminator $D_{\phi}$  via meta-learning \citep{finn2017maml}.
More concretely, the training phase of DCS is given as follows: for each training sample $\boldsymbol{x}_{tr}$, the latent optimization is carried out by minimizing $\left\| \boldsymbol{y}_{tr}- \boldsymbol{A}G_{\boldsymbol{\theta}}(\boldsymbol{z})\right\|$ while keeping $\boldsymbol{\theta}$ fixed, then $\boldsymbol{\theta}$ are subsequently trained in DCS: 
\begin{align}
    & \boldsymbol{z}^*_{\boldsymbol{x}_{tr}} = \underset{\boldsymbol{z}}{\arg\min} \left\| \boldsymbol{y}_{tr}- \boldsymbol{A}G_{\boldsymbol{\theta}}(\boldsymbol{z})\right\|,\label{meta_obj1} \\
    & \boldsymbol{\theta}^* = \mathcal{F}^{opt}_{\boldsymbol{\theta}}\Big[\mathcal{L}_{tr}\big(G_{\boldsymbol{\theta}}(\boldsymbol{z}^*_{\boldsymbol{x}_{tr}})\big)\Big] =  \underset{\boldsymbol{\theta}}{\arg\min}\, \mathbb{E}_{\boldsymbol{x}_{tr} \sim p(\boldsymbol{x})}\Big[\left\| \boldsymbol{y}_{tr}- \boldsymbol{A}G_{\boldsymbol{\theta}}(\boldsymbol{z}^*_{\boldsymbol{x}_{tr}})\right\|^2\Big]. \label{meta_obj2}
\end{align}
Note that the test phase of DCS is the same as that of CSGM \eqref{csgm}, so it is not shown here.

\paragraph{DCS-IM} Similarly to \eqref{im_dcgan_obj} and \eqref{im_began_obj}, the IM framework makes the measurement information also taken by $G_{\theta}$ as additional input in DCS. Therefore, by applying IM to DCS, \eqref{meta_obj1} and \eqref{meta_obj2}, the procedure for training $\boldsymbol{\theta}$, are substituted with \eqref{im_meta_obj1} and \eqref{im_meta_obj2}, respectively.
\begin{align}
    \boldsymbol{z}^*_{\boldsymbol{x}_{tr}} &= \underset{\boldsymbol{z}}{\arg\min} \left\| \boldsymbol{y}_{tr}- \boldsymbol{A}G_{\boldsymbol{\theta}}(\boldsymbol{z}, \begin{color}{black}\boldsymbol{y}_{tr}\end{color})\right\|,\label{im_meta_obj1} \\
    \boldsymbol{\theta}^*&= \mathcal{F}^{opt}_{\boldsymbol{\theta}}\Big[\mathcal{L}_{tr}\big(G_{\boldsymbol{\theta}}(\boldsymbol{z}^*_{\boldsymbol{x}_{tr}},\textcolor{black}{\boldsymbol{y}_{tr}})\big)\Big] =  \underset{\boldsymbol{\theta}}{\arg\min}\, \mathbb{E}_{\boldsymbol{x}_{tr} \sim p(\boldsymbol{x})}\Big[\left\| \boldsymbol{y}_{tr}- \boldsymbol{A}G_{\boldsymbol{\theta}}(\boldsymbol{z}^*_{\boldsymbol{x}_{tr}}, \begin{color}{black}\boldsymbol{y}_{tr}\end{color})\right\|^2\Big].\label{im_meta_obj2}
\end{align}

As the formula of DCS in the test phase is the same as that of CSGM \eqref{csgm}, the formula of DCS with the IM framework is also identical to that of CSGM-IM \eqref{csgm_im}, but notice that a couple of gradient descent steps are sufficient to implement \eqref{csgm} in DCS and \eqref{csgm_im} in DCS with the IM framework. 

In such a case of applying IM to DCS in the training and test phases, we name it DCS-IM. 
The algorithms of DCS and DCS-IM are given in Section \ref{sec_alg_scode}.

\subsection{Sparse deviations for compressed sensing using Generative models (SparseGen)}\label{subsec:sparse} 
\paragraph{SparseGen}  \citet{dhar2018modeling} came up with the idea to combine a domain-specific generative model prior with sparsity prior to enhance the generalization of CSGM, called SparseGen. 
Similarly to CSGM, the training phase of SparseGen can be any learning scheme to optimize a generator $G_{\boldsymbol{\theta}}$ like \eqref{dcgan_obj} or \eqref{began_obj}. In the test phase of SparseGen, given trained parameters $\boldsymbol{\theta}^*$ of a generator, sparse deviations from the support set of $G_{\boldsymbol{\theta}^*}$ are allowed to consider signals even outside the range of $G_{\boldsymbol{\theta}^*}$, which results in an estimate being of the form $G_{\boldsymbol{\theta}^*} + \boldsymbol{\nu}$ where $\boldsymbol{\nu}$ is an augmented sparse estimate. Hence, $\mathcal{L}_{te}$ should be involved with $\ell_1$ minimization with respect to $\boldsymbol{\nu}$ as well as the optimization of $G_{\boldsymbol{\theta}^*}$ with respect to $\boldsymbol{z}$:
\begin{align}\label{sparsegen}
  (\boldsymbol{z}^*, \boldsymbol{\nu}^*) = \mathcal{F}^{opt}_{(\boldsymbol{z}, \boldsymbol{\nu})}[\mathcal{L}_{te}(G_{\boldsymbol{\theta}^*}(\boldsymbol{z}), \boldsymbol{\nu} \, | \, \boldsymbol{y}_{te})]  = 
  \underset{\boldsymbol{z}, \boldsymbol{\nu}}{\arg\min}  \left\| \boldsymbol{A}(G_{{\boldsymbol{\theta}}^*}(\boldsymbol{z}) + \boldsymbol{\nu})-\boldsymbol{y}_{te} \right\| + \lambda \left\| \boldsymbol{B}\boldsymbol{\nu}\right\|_1,
\end{align}
where $\boldsymbol{B}$ is a transform matrix promoting sparsity of the vector $\boldsymbol{B}\boldsymbol{\nu}$, and $\lambda$ is the Lagrange multiplier. 
By using $(\boldsymbol{z}^*, \boldsymbol{\nu}^*)$ given in \eqref{sparsegen}, SparseGen estimates the target signal as $G_{{\boldsymbol{\theta}}^*}(\boldsymbol{z}^*)+\boldsymbol{\nu}^*$. 

\paragraph{SparseGen-IM}  IM can be run on SparseGen by adjusting \eqref{sparsegen} to \eqref{sparsegen-im}:
\begin{align}\label{sparsegen-im}
  (\boldsymbol{z}^*, \boldsymbol{\nu}^*) = \mathcal{F}^{opt}_{(\boldsymbol{z}, \boldsymbol{\nu})}[\mathcal{L}_{te}(G_{\boldsymbol{\theta}^*}(\boldsymbol{z}, \textcolor{black}{\boldsymbol{y}_{te}}), \boldsymbol{\nu} \, | \, \boldsymbol{y}_{te})]  = 
  \underset{\boldsymbol{z}, \boldsymbol{\nu}}{\arg\min}  \left\| \boldsymbol{A}(G_{{\boldsymbol{\theta}}^*}(\boldsymbol{z}, \textcolor{black}{\boldsymbol{y}_{te}}) + \boldsymbol{\nu})-\boldsymbol{y}_{te} \right\| + \lambda \left\| \boldsymbol{B}\boldsymbol{\nu}\right\|_1,
\end{align}
which we name SparseGen-IM. In this case, $G_{{\boldsymbol{\theta}}^*}(\boldsymbol{z}^*, \boldsymbol{y}_{te})+\boldsymbol{\nu}^*$ becomes an estimate of $\boldsymbol{x}_{te}$.  As SparseGen has the same training phase as CSGM, SparseGen-IM also has the same training phase as CSGM-IM (\eqref{im_dcgan_obj} or \eqref{im_began_obj}). 

The algorithms of SparseGen and SparseGen-IM are given in Section \ref{sec_alg_scode}.

\section{Proof of Theorem \ref{thm_im}}
The proof of Theorem \ref{thm_im} is based on that of Theorem 1 in \citet{jagatap2019algorithmic}.

Let the following algorithm be the test phase of SPGDGAN (without blue notes) and SPGDGAN-IM (with blue notes) in the noiseless setting, $\boldsymbol{y} = \boldsymbol{A}\boldsymbol{x}$. 
    \begin{algorithmic}[1]
    \Input{$\boldsymbol{y}_{te} \in \mathbb{R}^{m}, \boldsymbol{A} \in \mathbb{R}^{m \times d}$, $G_{{\boldsymbol{\theta}}^*}: \mathbb{R}^{v} \mapsto \mathbb{R}^{d}$, $\alpha \in \mathbb{R}_{+}$, $T \in  \mathbb{N}$}
    \Initialize{$\boldsymbol{x}_{0} = \boldsymbol{0} \in \mathbb{R}^{d}$} 
		\For{$t = 0$ to $T-1$} 
		\State $\boldsymbol{w}_t = \boldsymbol{x}_t - \alpha \boldsymbol{A}^{\top}(\boldsymbol{A}\boldsymbol{x}_t-\boldsymbol{y}_{te})$ 
		\State $ \boldsymbol{z}_t = \argmin\limits_{\boldsymbol{z}} \left\| \boldsymbol{w}_t - G_{\boldsymbol{\theta}^*}(\boldsymbol{z},\begin{color}{blue}\boldsymbol{y}_{te}\end{color})\right\|$
		\State $\boldsymbol{x}_{t+1} = \boldsymbol{U} h_s\Big(\boldsymbol{U}^{\top}   G_{\boldsymbol{\theta}^*}\big(\boldsymbol{z}_t, \begin{color}{blue}\boldsymbol{y}_{te}\end{color} \big)\Big)$  
		\EndFor
    \Output{the signal estimate $\hat{\boldsymbol{x}} = \boldsymbol{x}_{T}$}
  \end{algorithmic}
Referring to the above algorithm, we prove Theorem  \ref{thm_im} under the condition (b) (i.e., the guarantee of signal recovery in SPGDGAN-IM). The proof of Theorem  \ref{thm_im} under the condition (a)  (i.e., the guarantee of signal recovery in SPGDGAN) is trivial
if we remove all the blue notes below and use the condition (a) instead of (b).  

It follows that 
\begin{align}\nonumber
&\left\|\boldsymbol{y}_{te}-\boldsymbol{A}\boldsymbol{x}_{t+1} \right\|^2 - \left\|\boldsymbol{y}_{te}-\boldsymbol{A}\boldsymbol{x}_{t} \right\|^2  \\\nonumber
&= (\left\| \boldsymbol{Ax}_{t+1} \right\|^2-\left\| \boldsymbol{Ax}_{t} \right\|^2) - 2 (\boldsymbol{y}_{te}^{\top}\boldsymbol{Ax}_{t+1}-\boldsymbol{y}_{te}^{\top}\boldsymbol{Ax}_{t})\\\nonumber
&= \left\| \boldsymbol{A}\boldsymbol{x}_{t+1} - \boldsymbol{A}\boldsymbol{x}_{t} \right\|^2  -2(\boldsymbol{A}\boldsymbol{x}_{t})^{\top}(\boldsymbol{A}\boldsymbol{x}_{t}) +2(\boldsymbol{A}\boldsymbol{x}_{t})^{\top}(\boldsymbol{A}\boldsymbol{x}_{t+1}) - 2 (\boldsymbol{y}_{te}^{\top}\boldsymbol{A}\boldsymbol{x}_{t+1}-\boldsymbol{y}_{te}^{\top}\boldsymbol{A}\boldsymbol{x}_{t}) \\\label{eq_temp1}
&= \left\| \boldsymbol{A}\boldsymbol{x}_{t+1} - \boldsymbol{A}\boldsymbol{x}_{t} \right\|^2  + 2(\boldsymbol{A}\boldsymbol{x}_{t}-\boldsymbol{y}_{te})^{\top} (\boldsymbol{A}\boldsymbol{x}_{t+1}-\boldsymbol{A}\boldsymbol{x}_{t}).
\end{align}

Line 2 of the SPGDGAN-IM algorithm is given as 
\begin{align}\label{line2}
\boldsymbol{w}_t = \boldsymbol{x}_t - \alpha \boldsymbol{A}^{\top}(\boldsymbol{A}\boldsymbol{x}_t-\boldsymbol{y}_{te}). 
\end{align}
Afterward, by using $\boldsymbol{w}_t$ in  \eqref{line2}, line 3 of the SPGDGAN-IM algorithm updates the latent variables $\boldsymbol{z}$ so that  $G_{\boldsymbol{\theta}^*}(\boldsymbol{z},\begin{color}{blue}\boldsymbol{y}_{te}\end{color})$ lies in the range of the generator $G_{\boldsymbol{\theta}^*}$ while being the closest to $\boldsymbol{w}_t$.

From line 4 of the SPGDGAN algorithm, $\boldsymbol{U} h_s\Big(\boldsymbol{U}^{\top}\big( G_{\boldsymbol{\theta}^*}(\boldsymbol{z}_t, \textcolor{blue}{\boldsymbol{y}_{te}})\big)\Big)=\boldsymbol{x}_{t+1}$. By the condition (b), there exists $\boldsymbol{z}^*$ satisfying $ G_{\boldsymbol{\theta}^*}(\boldsymbol{z}^*, \textcolor{blue}{\boldsymbol{y}_{te}})=\boldsymbol{x}_{te}$ so that $\boldsymbol{U} h_s\Big(\boldsymbol{U}^{\top}\big( G_{\boldsymbol{\theta}^*}(\boldsymbol{z}^*, \textcolor{blue}{\boldsymbol{y}_{te}})\big)\Big)=\boldsymbol{x}_{te}$. Then, by using these two results and the definition of $\boldsymbol{z}_{t}$ in line 3, we obtain  
\begin{align}\label{line3}
\left\| \boldsymbol{x}_{t+1} -  \boldsymbol{w}_{t} \right\|^2 \leq \left\| \boldsymbol{x}_{te} -  \boldsymbol{w}_{t} \right\|^2.
\end{align}
By applying \eqref{line2} to \eqref{line3}, we obtain  
\begin{align}\label{line3_1}
\left\| \boldsymbol{x}_{t+1} -  \boldsymbol{x}_t + \alpha \boldsymbol{A}^{\top}(\boldsymbol{A}\boldsymbol{x}_t-\boldsymbol{y}_{te}) \right\|^2 \leq \left\| \boldsymbol{x}_{te} -  \boldsymbol{x}_t + \alpha \boldsymbol{A}^{\top}(\boldsymbol{A}\boldsymbol{x}_t-\boldsymbol{y}_{te})\right\|^2. 
\end{align}
\eqref{line3_1} can be written as 
\begin{align*}
\left\| \boldsymbol{x}_{t+1} - \boldsymbol{x}_{t} \right\|^2 + 2\alpha\left\|(\boldsymbol{x}_{t+1} - \boldsymbol{x}_t)^T\boldsymbol{A}^T(\boldsymbol{A}\boldsymbol{x}_t - \boldsymbol{y}_{te})\right\| &\le \left\| \boldsymbol{x}_{te} - \boldsymbol{x}_{t} \right\|^2 + 2\alpha\left\|(\boldsymbol{x}_{te}-\boldsymbol{x}_t)^T\boldsymbol{A}^T(\boldsymbol{A}\boldsymbol{x}_t-\boldsymbol{y}_{te})\right\| \\
&= \left\| \boldsymbol{x}_{te} - \boldsymbol{x}_{t} \right\|^2 + 2\alpha\left\|(\boldsymbol{A}\boldsymbol{x}_{te}-\boldsymbol{A}\boldsymbol{x}_t)^T(\boldsymbol{A}\boldsymbol{x}_t-\boldsymbol{y}_{te})\right\| \\
&= \left\| \boldsymbol{x}_{te} - \boldsymbol{x}_{t} \right\|^2 + 2\alpha\left\|(\boldsymbol{y}_{te}-\boldsymbol{A}\boldsymbol{x}_t)^T(\boldsymbol{A}\boldsymbol{x}_t-\boldsymbol{y}_{te})\right\| \\
&= \left\| \boldsymbol{x}_{te} - \boldsymbol{x}_{t} \right\|^2 -2\alpha\left\|\boldsymbol{y}_{te}-\boldsymbol{A}\boldsymbol{x}_t\right\|^2.
\end{align*}
By dividing both sides by $\alpha$,
\begin{align}
\frac{1}{\alpha}\left\| \boldsymbol{x}_{t+1} - \boldsymbol{x}_{t} \right\|^2  + 2 (\boldsymbol{A}\boldsymbol{x}_t-\boldsymbol{y}_{te})^{\top}\boldsymbol{A}(\boldsymbol{x}_{t+1}-\boldsymbol{x}_t) \leq \frac{1}{\alpha} \left\| \boldsymbol{x}_{t} - \boldsymbol{x}_{te}\right\|^2 - 2 \left\|\boldsymbol{y}_{te}-\boldsymbol{A}\boldsymbol{x}_{t} \right\|^2.\label{line3_2}
\end{align}
Then, by applying \eqref{eq_temp1} to \eqref{line3_2}, we gain 
\begin{align}\label{line3_3} 
\left\|\boldsymbol{y}_{te}-\boldsymbol{A}\boldsymbol{x}_{t+1} \right\|^2 +  \left\|\boldsymbol{y}_{te}-\boldsymbol{A}\boldsymbol{x}_{t} \right\|^2 \leq \frac{1}{\alpha} \left\| \boldsymbol{x}_{t} - \boldsymbol{x}_{te}\right\|^2 - \frac{1}{\alpha} \left\| \boldsymbol{x}_{t+1} - \boldsymbol{x}_{t}\right\|^2 + \left\|  \boldsymbol{A}( \boldsymbol{x}_{t+1} - \boldsymbol{x}_{t})\right\|^2.
\end{align}
From $(\mathcal{S},1-\gamma,1+\gamma)$-RIP condition, the first and last terms on the right-hand side of \eqref{line3_3} are upper bounded respectively by 
\begin{align}\nonumber
\frac{1}{\alpha} \left\| \boldsymbol{x}_{t} - \boldsymbol{x}_{te}\right\|^2 
&= \frac{1}{\alpha} \left\| \boldsymbol{U} h_s(\boldsymbol{U}^{\top}\boldsymbol{x}_t) - \boldsymbol{U} h_s(\boldsymbol{U}^{\top}\boldsymbol{x}_{te}) \right\|^2  \\\nonumber
&= \frac{1}{\alpha} \left\|  h_s(\boldsymbol{U}^{\top}\boldsymbol{x}_t) -  h_s(\boldsymbol{U}^{\top}\boldsymbol{x}_{te}) \right\|^2  \\\nonumber
&\overset{(a)}{\leq} \frac{1}{\alpha (1- \gamma)} \left\|\boldsymbol{A}\boldsymbol{U}(h_s(\boldsymbol{U}^{\top}\boldsymbol{x}_t) -  h_s(\boldsymbol{U}^{\top}\boldsymbol{x}_{te}))\right\|^2 \\\label{bound1} 
&\overset{(b)}{=} \frac{1}{\alpha (1- \gamma)} \left\|\boldsymbol{A}(\boldsymbol{x}_{t} - \boldsymbol{x}_{te})\right\|^2 
\end{align}
and 
\begin{align}\nonumber
\left\|  \boldsymbol{A}( \boldsymbol{x}_{t+1} - \boldsymbol{x}_{t})\right\|^2 
&= \left\|\boldsymbol{A}\boldsymbol{U}(h_s(\boldsymbol{U}^{\top}\boldsymbol{x}_{t+1}) -  h_s(\boldsymbol{U}^{\top}\boldsymbol{x}_{t}))\right\|^2 \\\nonumber 
&\overset{(c)}{\leq} (1+\gamma) \left\|  h_s(\boldsymbol{U}^{\top}\boldsymbol{x}_{t+1}) -  h_s(\boldsymbol{U}^{\top}\boldsymbol{x}_{t}) \right\|^2\\\nonumber 
&\overset{(d)}{=} (1+\gamma) \left\|  \boldsymbol{U}^{\top}\boldsymbol{x}_{t+1} -  \boldsymbol{U}^{\top}\boldsymbol{x}_{t} \right\|^2
\\\label{bound2} 
&= (1+\gamma) \left\|  \boldsymbol{x}_{t+1} - \boldsymbol{x}_{t}\right\|^2,
\end{align}
where (c) follows from RIP and (d) follows from $\boldsymbol{x}_t =  \boldsymbol{U} h_s(\boldsymbol{U}^{\top}\boldsymbol{x}_t)$. \\
Then, it follows that  
\begin{align}\nonumber 
\left\|\boldsymbol{y}_{te}-\boldsymbol{A}\boldsymbol{x}_{t+1} \right\|^2 +  \left\|\boldsymbol{y}_{te}-\boldsymbol{A}\boldsymbol{x}_{t} \right\|^2 
&\overset{}{\leq} \frac{1}{\alpha} \left\| \boldsymbol{x}_{t} - \boldsymbol{x}_{te}\right\|^2 - \frac{1}{\alpha} \left\| \boldsymbol{x}_{t+1} - \boldsymbol{x}_{t}\right\|^2 + \left\|  \boldsymbol{A}( \boldsymbol{x}_{t+1} - \boldsymbol{x}_{t})\right\|^2 \\\nonumber
&\overset{(e)}{\leq}
\frac{1}{\alpha (1- \gamma)}  \left\|\boldsymbol{y}_{te}-\boldsymbol{A}\boldsymbol{x}_{t} \right\|^2  + (1+\gamma - \frac{1}{\alpha}) \left\|  \boldsymbol{x}_{t+1} -\boldsymbol{x}_{t}\right\|^2  \\\label{line3_4}
&\overset{(f)}{\leq} 
\frac{1}{\alpha (1- \gamma)}  \left\|\boldsymbol{y}_{te}-\boldsymbol{A}\boldsymbol{x}_{t} \right\|^2,  
\end{align}
where (e) is satisfied by applying \eqref{bound1} and  \eqref{bound2} to \eqref{line3_3}, and (f) follows from a supplementary assumption that $\alpha < 1/(1+\gamma)$.

By moving the second term on the left-hand side of \eqref{line3_4} to the right-hand side,
\begin{align}\label{conv_formular1} 
\left\|\boldsymbol{y}_{te}-\boldsymbol{A}\boldsymbol{x}_{t+1} \right\|^2    {\leq} 
  \Big(\frac{1}{\alpha (1- \gamma)} -1 \Big) \left\|\boldsymbol{y}_{te}-\boldsymbol{A}\boldsymbol{x}_{t} \right\|^2.
\end{align}
By repeating \eqref{conv_formular1} for $t \in \{0,1,...,T-1\}$, we get 
\begin{align}\label{conv_formular2} 
\left\|\boldsymbol{y}_{te}-\boldsymbol{A}\boldsymbol{x}_{T} \right\|^2    {\leq} 
  \Big(\frac{1}{\alpha (1- \gamma)} -1 \Big)^T \left\|\boldsymbol{y}_{te}-\boldsymbol{A}\boldsymbol{x}_{0} \right\|^2.
\end{align}
From \eqref{bound1}, we also get 
\begin{align}\label{conv_formular3} 
\left\|\boldsymbol{x}_{te}-\boldsymbol{x}_{T} \right\|^2  \leq \frac{1}{1-\gamma}\left\|\boldsymbol{y}_{te}-\boldsymbol{A}\boldsymbol{x}_{T} \right\|^2.
\end{align}
Applying \eqref{conv_formular2} to \eqref{conv_formular3}, we obtain
\begin{align}\label{conv_formular4} 
\left\|\boldsymbol{x}_{te}-\boldsymbol{x}_{T} \right\|^2  \leq \frac{1}{1-\gamma} \Big(\frac{1}{\alpha (1- \gamma)} -1 \Big)^T \left\|\boldsymbol{y}_{te}-\boldsymbol{A}\boldsymbol{x}_{0} \right\|^2 \triangleq \epsilon^2.
\end{align}
If we set $\alpha$ to any constant satisfying $\frac{1}{2(1-\gamma)} < \alpha < \frac{1}{(1-\gamma)}$ and $\alpha < \frac{1}{(1+\gamma)}$,
the right-hand side of \eqref{conv_formular4} converges to zero when $T$ is sufficiently large. 
In other words, if we denote $v:=(\frac{1}{\alpha (1- \gamma)} -1 )^{-1}$, 
and $T$ satisfies 
\begin{align}\label{conv_formular5} 
T = \log_{v} \Bigg(\frac{\left\|\boldsymbol{y}_{te}-\boldsymbol{A}\boldsymbol{x}_{0} \right\|^2}{\epsilon^2 (1-\gamma)}\Bigg) \propto \log \Big(\frac{1}{\epsilon}\Big),
\end{align}
then we obtain the following equation, thereby completing the proof.
\begin{align}\label{conv_formular6} 
\left\|\boldsymbol{x}_{te}-\boldsymbol{x}_{T} \right\|  \leq \epsilon.
\end{align}

\section{Proof of Theorem \ref{thm2}}
Our proof is based on the following theorem by Davidson and Szarek \cite{davidson2001local}.
\begin{thm}[$\textup{\cite[Theorem II.13]{davidson2001local}}$]: Given $k,m,d \in \mathbb{N}$ satisfying $k \leq m \leq d$, consider the random matrix $\boldsymbol{A} \in \mathbb{R}^{m \times d}$ whose entries
are i.i.d. Gaussian following $\mathcal{N}(0,1/m)$. Then, for any $t >0$ and any set $\Gamma \in \{d\}$ satisfying $|\Gamma|=k$, 
\begin{align}\nonumber
    \mathbb{P}\Big(\sigma_{1}(\boldsymbol{A}_{\Gamma}) \geq 1+\sqrt{\frac{k}{m}}+t\Big) \leq \exp\Big(- \frac{mt^2}{2}\Big), \,\,\,
    \mathbb{P}\Big(\sigma_{m}(\boldsymbol{A}_{\Gamma}) \leq 1-\sqrt{\frac{k}{m}}-t\Big) \leq \exp\Big(- \frac{mt^2}{2}\Big).
\end{align}
\label{thm_gas}
\end{thm}
Note that the condition for $m$ in Theorem \ref{thm2} implies $1-\sqrt{1-\gamma}\geq \sqrt{1+\gamma}-1 \geq 2 \sqrt{\frac{k}{m}}$ for $k \leq 2s$. By using these inequalities, it follows that Theorem \ref{thm_gas} implies both of the following inequalities hold.
\begin{align}\nonumber
    &\mathbb{P}\Big(\sigma_{1}(\boldsymbol{A}_{\Gamma}) \geq  \sqrt{1+\gamma}\Big) \leq \exp\Bigg(- \frac{m}{2}\Big(\sqrt{1+\gamma}-1 -\sqrt{\frac{k}{m}}\Big)^2\Bigg) \\\nonumber
    &\mathbb{P}\Big(\sigma_{m}(\boldsymbol{A}_{\Gamma}) \leq \sqrt{1-\gamma}\Big) \leq \exp\Bigg(- \frac{m}{2}\Big(1-\sqrt{1-\gamma}-\sqrt{\frac{k}{m}}\Big)^2\Bigg)
\end{align}
As the condition for $m$ in Theorem \ref{thm2} implies $ 1-\sqrt{1-\gamma}-\sqrt{\frac{k}{m}} \geq   \sqrt{1+\gamma}-1 -\sqrt{\frac{k}{m}} > 0$, we get
\begin{align}\label{bound_rip1}
\mathbb{P}\Big(\max\Big[\sigma^2_{1}(\boldsymbol{A}_{\Gamma})-1,1-\sigma^2_{m}(\boldsymbol{A}_{\Gamma})\Big]\geq \gamma \Big) \leq 2 \exp \Big( -\frac{m}{2} \big( \sqrt{1+\gamma}-1 -\sqrt{\frac{k}{m}} \big)^2 \Big). 
\end{align}
Due to the rotational invariance of Gaussian vector, each element of $\boldsymbol{A}\boldsymbol{U}$ independently follows $\mathcal{N}(0,1/m)$ as well as $\boldsymbol{A}$, where $\boldsymbol{U}$ is an arbitrary unitary matrix. Therefore, $\boldsymbol{A}$ in \eqref{bound_rip1} can be replaced by $\boldsymbol{A}\boldsymbol{U}$ as follows.
\begin{align}\label{bound_rip2}
\mathbb{P}\Big(\max\Big[\sigma^2_{1}((\boldsymbol{A}\boldsymbol{U})_{\Gamma})-1,1-\sigma^2_{m}((\boldsymbol{A}\boldsymbol{U})_{\Gamma})\Big]\geq \gamma \Big) \leq 2 \exp \Big( -\frac{m}{2} \big( \sqrt{1+\gamma}-1 -\sqrt{\frac{k}{m}} \big)^2 \Big). 
\end{align} 
By using the union of events $\Big(\max\Big[\sigma^2_{1}((\boldsymbol{A}\boldsymbol{U})_{\Gamma})-1,1-\sigma^2_{m}((\boldsymbol{A}\boldsymbol{U})_{\Gamma})\Big]\geq \gamma \Big)$ for every different $\Gamma$ belonging to the set $\Sigma$ defined in Theorem \ref{thm2} (i.e., at most $2T$ times), we get 
\begin{align}\nonumber
\mathbb{P}\Big(\textup{$\boldsymbol{AU}$ does not satisfy $(\mathcal{S},1-\gamma,1+\gamma)$-RIP}\Big)
&=\mathbb{P}\Big(\max_{\Gamma \in \Sigma} \Big[\max\Big[\sigma^2_{1}((\boldsymbol{A}\boldsymbol{U})_{\Gamma})-1,1-\sigma^2_{m}((\boldsymbol{A}\boldsymbol{U})_{\Gamma})\Big]\Big]\geq \gamma \Big)\\\label{temp1}
&\leq 4 T \exp \Big( -\frac{m}{2} \big( \sqrt{1+\gamma}-1 -\sqrt{\frac{2s}{m}} \big)^2 \Big),
\end{align}
where we set $k=2s$ as each set in $\Sigma$ has the sparsity of at most $2s$. 
The RHS of \eqref{temp1} is upper bounded by $\tau$ if $(\sqrt{1+\gamma}-1)\sqrt{m} \geq \sqrt{2s} + \sqrt{2\ln(\frac{4T}{\tau})}$, which is implied by $(\sqrt{1+\gamma}-1)\sqrt{m} \geq \sqrt{2\Big(s+  \ln(\frac{4T}{\tau})\Big)}$.

\clearpage
\section{Algorithm Pseudocode}\label{sec_alg_scode}
\begin{algorithm}
  \caption{CSGM (without blue notes) and CSGM-IM in the case of using DCGAN training objective}
    \begin{algorithmic}[H]
    \TR
    \Input{$\boldsymbol{A} \in \mathbb{R}^{m \times d}$, $G_{\boldsymbol{\theta}}: \mathbb{R}^{v} \mapsto \mathbb{R}^{d}$, $ D_{\boldsymbol{\phi}}: \mathbb{R}^{d} \mapsto [0, 1]$}
    	\For{$i = 1$ to $n$}
    	\State sample $\boldsymbol{x}_{tr,i}$ from $p(\boldsymbol{x})$  
    	\State measure $\boldsymbol{y}_{tr,i}=\boldsymbol{A} \boldsymbol{x}_{tr,i}$
    	\State sample $\boldsymbol{z}_{i}$ from $p_{\boldsymbol{z}}(\boldsymbol{z})$  
    	\EndFor
    	\Comment{generate $n$ training samples}
    	\vskip +1pt
    	\State $({\boldsymbol{\theta}}^*,{\boldsymbol{\phi}}^*)=\underset{\boldsymbol{\theta}}{\argmin}\, \underset{\boldsymbol{\phi}}{\argmax}\,\frac{1}{n}\sum\limits_{i=1}^n [\ln \, D_{\boldsymbol{\phi}}(\boldsymbol{x}_{tr,i}, \begin{color}{blue}\boldsymbol{y}_{tr,i}\end{color}) + \ln \, (1-D_{\boldsymbol{\phi}}(G_{\boldsymbol{\theta}}(\boldsymbol{z}_i, \begin{color}{blue}\boldsymbol{y}_{tr,i}\end{color}), \begin{color}{blue}\boldsymbol{y}_{tr,i}\end{color}))]$ 
    \Output{the trained parameters ${\boldsymbol{\theta}}^*$ of $G$}
  \end{algorithmic}
  \algrule
    \begin{algorithmic}[H]
    \TE
    \Input{$\boldsymbol{y}_{te} \in \mathbb{R}^{m}, \boldsymbol{A} \in \mathbb{R}^{m \times d}$, $G_{\boldsymbol{\theta}^*}: \mathbb{R}^{v} \mapsto \mathbb{R}^{d}$, $\tau  \in \mathbb{R}_{+}$, $T\in \mathbb{N}$}
    \Initialize{sample $\boldsymbol{z}_0$ from $p_{\boldsymbol{z}}(\boldsymbol{z})$}
	\For{$t = 0$ to $T-1$}
	\State $\boldsymbol{z}_{t+1} = \boldsymbol{z}_{t} - \tau \frac{\partial}{\partial \boldsymbol{z}}\left\| \boldsymbol{y}_{te} - \boldsymbol{A}G_{\boldsymbol{\theta}^*}(\boldsymbol{z}, \begin{color}{blue}\boldsymbol{y}_{te}\end{color}) \right\|^2\Big|_{\boldsymbol{z}=\boldsymbol{z}_t}$
	\EndFor
    \Output{the signal estimate $\hat{\boldsymbol{x}} = G_{\boldsymbol{\theta}^*}(\boldsymbol{z}_{T}, \begin{color}{blue}\boldsymbol{y}_{te}\end{color})$}
  \end{algorithmic}
  \label{alg_csgm_dcgan}
\end{algorithm}

\begin{algorithm}
  \caption{CSGM (without blue notes) and CSGM-IM in the case of using BEGAN training objective}
    \begin{algorithmic}[H]
    \TR
    \Input{$\boldsymbol{A} \in \mathbb{R}^{m \times d}$, $G_{\boldsymbol{\theta}}: \mathbb{R}^{v} \mapsto \mathbb{R}^{d}$, $D_{\boldsymbol{\phi}}: \mathbb{R}^{d} \mapsto \mathbb{R}^{d}$, $R_{\boldsymbol{\phi}}(\bar{\boldsymbol{x}}, \textcolor{blue}{\boldsymbol{y}}) = |\bar{\boldsymbol{x}} - D_{\boldsymbol{\phi}}(\bar{\boldsymbol{x}}, \textcolor{blue}{\boldsymbol{y}})|$, $\lambda \in \mathbb{R}_{+}$, $\gamma \in [0, 1]$, $\zeta(0) = 0$, $K\in \mathbb{N}$}
    	\For{$i = 1$ to $n$}
    	\State sample $\boldsymbol{x}_{tr,i}$ from $p(\boldsymbol{x})$  
    	\State measure $\boldsymbol{y}_{tr,i}=\boldsymbol{A} \boldsymbol{x}_{tr,i}$
    	\State sample $\boldsymbol{z}^{G}_{i}$ and $\boldsymbol{z}^{D}_{i}$ independently from $p_{\boldsymbol{z}}(\boldsymbol{z})$  
    	\EndFor
    	\Comment{generate $n$ training samples}
    	\vskip +1pt
    	\For{$k = 0$ to $K-1$} 
    	\State $\boldsymbol{\phi}(k+1) = \boldsymbol{\phi}(k) - \eta \frac{\partial}{\partial \boldsymbol{\phi}}\Big( \sum\limits_{i=1}^n \big( R_{\boldsymbol{\phi}}(\boldsymbol{x}_{tr,i}, \begin{color}{blue}\boldsymbol{y}_{tr,i}\end{color}) - \zeta(k) R_{\boldsymbol{\phi}}(G_{\boldsymbol{\theta}(k)}(\boldsymbol{z}^{D}_{i}, \begin{color}{blue}\boldsymbol{y}_{tr,i}\end{color}), \begin{color}{blue}\boldsymbol{y}_{tr,i}\end{color}) \big) \Big) \Big|_{\boldsymbol{\phi}=\boldsymbol{\phi}(k)}$
    	\State $\boldsymbol{\theta}(k+1) = \boldsymbol{\theta}(k) - \eta \frac{\partial}{\partial \boldsymbol{\theta}}\Big(\sum\limits_{i=1}^n R_{\boldsymbol{\phi}(k)}(G_{\boldsymbol{\theta}}(\boldsymbol{z}^{G}_{i}, \begin{color}{blue}\boldsymbol{y}_{tr,i}\end{color}), \begin{color}{blue}\boldsymbol{y}_{tr,i}\end{color}) \Big)   \Big|_{\boldsymbol{\theta}=\boldsymbol{\theta}(k)}$
        \State  $\zeta(k+1)=\min(\max(\zeta(k)+\lambda (\gamma R_{\boldsymbol{\phi}(k)}(\boldsymbol{x}_{tr,i}, \begin{color}{blue}\boldsymbol{y}_{tr,i}\end{color})-R_{\boldsymbol{\phi}(k)}(G_{\boldsymbol{\theta}(k)}(\boldsymbol{z}^{G}_{i}, \begin{color}{blue}\boldsymbol{y}_{tr,i}\end{color}), \begin{color}{blue}\boldsymbol{y}_{tr,i}\end{color})),0),1)$
        \vskip +5pt
        \EndFor \Comment{optimize the network parameters}
    \Output{the trained parameter ${\boldsymbol{\theta}}(K)$ of $G$}
  \end{algorithmic}
  \algrule
    \begin{algorithmic}[H]
    \TE
    \Input{$\boldsymbol{y}_{te} \in \mathbb{R}^{m}, \boldsymbol{A} \in \mathbb{R}^{m \times d}$, $G_{\boldsymbol{\theta}({K})}: \mathbb{R}^{v} \mapsto \mathbb{R}^{d}$, $\tau  \in \mathbb{R}_{+}$, $T\in \mathbb{N}$}
    \Initialize{sample $\boldsymbol{z}_0$ from $p_{\boldsymbol{z}}(\boldsymbol{z})$}
	\For{$t = 0$ to $T-1$}
	\State $\boldsymbol{z}_{t+1} = \boldsymbol{z}_{t} - \tau \frac{\partial}{\partial \boldsymbol{z}}\left\| \boldsymbol{y}_{te} - \boldsymbol{A}G_{\boldsymbol{\theta}({K})}(\boldsymbol{z}, \begin{color}{blue}\boldsymbol{y}_{te}\end{color}) \right\|^2\Big|_{\boldsymbol{z}=\boldsymbol{z}_t}$
	\EndFor
    \Output{the signal estimate $\hat{\boldsymbol{x}} = G_{\boldsymbol{\theta}({K})}(\boldsymbol{z}_{T}, \begin{color}{blue}\boldsymbol{y}_{te}\end{color})$}
  \end{algorithmic}
  \label{alg_csgm_began}
\end{algorithm}

\begin{algorithm}
  \caption{PGDGAN (without blue notes) and PGDGAN-IM in the case of using DCGAN training objective}
    \begin{algorithmic}[H]
    \TR: 
    \Input{$\boldsymbol{A} \in \mathbb{R}^{m \times d}$, $G_{\boldsymbol{\theta}}: \mathbb{R}^{v} \mapsto \mathbb{R}^{d}$, $ D_{\boldsymbol{\phi}}: \mathbb{R}^{d} \mapsto [0, 1]$}
    	\For{$i = 1$ to $n$}
    	\State sample $\boldsymbol{x}_{tr,i}$ from $p(\boldsymbol{x})$  
    	\State measure $\boldsymbol{y}_{tr,i}=\boldsymbol{A} \boldsymbol{x}_{tr,i}$
    	\State sample $\boldsymbol{z}_{i}$ from $p_{\boldsymbol{z}}(\boldsymbol{z})$  
    	\EndFor
    	\Comment{generate $n$ training samples}
    	\vskip +1pt
    	\State $({\boldsymbol{\theta}}^*,{\boldsymbol{\phi}}^*)=\underset{\boldsymbol{\theta}}{\argmin}\, \underset{\boldsymbol{\phi}}{\argmax}\,\frac{1}{n}\sum\limits_{i=1}^n [\ln \, D_{\boldsymbol{\phi}}(\boldsymbol{x}_{tr,i}, \begin{color}{blue}\boldsymbol{y}_{tr,i}\end{color}) + \ln \, (1-D_{\boldsymbol{\phi}}(G_{\boldsymbol{\theta}}(\boldsymbol{z}_i, \begin{color}{blue}\boldsymbol{y}_{tr,i}\end{color}), \begin{color}{blue}\boldsymbol{y}_{tr,i}\end{color}))]$ 
    \Output{the trained parameters ${\boldsymbol{\theta}}^*$ of $G$}
  \end{algorithmic}
  \algrule
    \begin{algorithmic}[H]
    \TE
    \Input{$\boldsymbol{y}_{te} \in \mathbb{R}^{m}, \boldsymbol{A} \in \mathbb{R}^{m \times d}$, $G_{{\boldsymbol{\theta}}^*}: \mathbb{R}^{v} \mapsto \mathbb{R}^{d}$, $(\alpha, \tau) \in \mathbb{R}^2_{+}$, $(T, K) \in  \mathbb{N}^2$}
    \Initialize{$\boldsymbol{x}_{0} = \boldsymbol{0} \in \mathbb{R}^{d}$} 
		\For{$t = 0$ to $T-1$}
		\State $\boldsymbol{w}_{t} = \boldsymbol{x}_{t} -\alpha \boldsymbol{A}^{\top}(\boldsymbol{A} \boldsymbol{x}_{t} - \boldsymbol{y}_{te})$
	    \State sample $\boldsymbol{z}_{0}$ from $p_{\boldsymbol{z}}(\boldsymbol{z})$
		\For{$k = 0$ to $K-1$}
        \State $\boldsymbol{z}_{k+1} = \boldsymbol{z}_{k} - \tau \frac{\partial}{\partial \boldsymbol{z}} \left\| \boldsymbol{w}_t - G_{\boldsymbol{\theta}^*}(\boldsymbol{z}, \begin{color}{blue}\boldsymbol{y}_{te}\end{color}) \right\|^2\Big|_{\boldsymbol{z}=\boldsymbol{z}_k}$ 
        \EndFor
		\State $\boldsymbol{x}_{t+1} = G_{\boldsymbol{\theta}^*}\big(\boldsymbol{z}_K, \begin{color}{blue}\boldsymbol{y}_{te}\end{color} \big)$ 
		\EndFor
    \Output{the signal estimate $\hat{\boldsymbol{x}} = \boldsymbol{x}_{T}$}
  \end{algorithmic}
  \label{alg_pgd}
\end{algorithm}

\begin{algorithm}
  \caption{SPGDGAN (without blue notes) and SPGDGAN-IM in the case of using DCGAN training objective}
    \begin{algorithmic}[H]
    \TR:
    \Input{$\boldsymbol{A} \in \mathbb{R}^{m \times d}$, $G_{\boldsymbol{\theta}}: \mathbb{R}^{v} \mapsto \mathbb{R}^{d}$, $ D_{\boldsymbol{\phi}}: \mathbb{R}^{d} \mapsto [0, 1]$}
    	\For{$i = 1$ to $n$}
    	\State sample $\boldsymbol{x}_{tr,i}$ from $p(\boldsymbol{x})$  
    	\State measure $\boldsymbol{y}_{tr,i}=\boldsymbol{A} \boldsymbol{x}_{tr,i}$
    	\State sample $\boldsymbol{z}_{i}$ from $p_{\boldsymbol{z}}(\boldsymbol{z})$  
    	\EndFor
    	\Comment{generate $n$ training samples}
    	\vskip +1pt
    	\State $({\boldsymbol{\theta}}^*,{\boldsymbol{\phi}}^*)=\underset{\boldsymbol{\theta}}{\argmin}\, \underset{\boldsymbol{\phi}}{\argmax}\,\frac{1}{n}\sum\limits_{i=1}^n [\ln \, D_{\boldsymbol{\phi}}(\boldsymbol{x}_{tr,i}, \begin{color}{blue}\boldsymbol{y}_{tr,i}\end{color}) + \ln \, (1-D_{\boldsymbol{\phi}}(G_{\boldsymbol{\theta}}(\boldsymbol{z}_i, \begin{color}{blue}\boldsymbol{y}_{tr,i}\end{color}), \begin{color}{blue}\boldsymbol{y}_{tr,i}\end{color}))]$ 
    \Output{the trained parameters ${\boldsymbol{\theta}}^*$ of $G$}
  \end{algorithmic}
  \algrule
    \begin{algorithmic}[H]
    \TE
    \Input{$\boldsymbol{y}_{te} \in \mathbb{R}^{m}, \boldsymbol{A} \in \mathbb{R}^{m \times d}$, $G_{{\boldsymbol{\theta}}^*}: \mathbb{R}^{v} \mapsto \mathbb{R}^{d}$, $(\alpha, \tau) \in \mathbb{R}^2_{+}$, $(T, K) \in  \mathbb{N}^2, \boldsymbol{U} \in \mathbb{R}^{d \times d}$}
    \Initialize{$\boldsymbol{x}_{0} = \boldsymbol{0} \in \mathbb{R}^{d}, s = d/2, h_s: \mathbb{R}^d \mapsto \mathbb{R}^d$} 
		\For{$t = 0$ to $T-1$}
		\State $\boldsymbol{w}_{t} = \boldsymbol{x}_{t} -\alpha \boldsymbol{A}^{\top}(\boldsymbol{A} \boldsymbol{x}_{t} - \boldsymbol{y}_{te})$
	    \State sample $\boldsymbol{z}_{0}$ from $p_{\boldsymbol{z}}(\boldsymbol{z})$
		\For{$k = 0$ to $K-1$}
        \State $\boldsymbol{z}_{k+1} = \boldsymbol{z}_{k} - \tau \frac{\partial}{\partial \boldsymbol{z}} \left\| \boldsymbol{w}_t - G_{\boldsymbol{\theta}^*}(\boldsymbol{z}, \begin{color}{blue}\boldsymbol{y}_{te}\end{color}) \right\|^2\Big|_{\boldsymbol{z}=\boldsymbol{z}_k}$ 
        \EndFor
		\State $\boldsymbol{x}_{t+1} = \boldsymbol{U}h_s\big(\boldsymbol{U}^TG_{\boldsymbol{\theta}^*}\big(\boldsymbol{z}_K, \begin{color}{blue}\boldsymbol{y}_{te}\end{color} \big) \big)$ 
		\EndFor
    \Output{the signal estimate $\hat{\boldsymbol{x}} = \boldsymbol{x}_{T}$}
  \end{algorithmic}
  \label{alg_spgd}
\end{algorithm}

\begin{algorithm}
  \caption{DCS (without blue notes) and DCS-IM}
    \begin{algorithmic}[H]
    \TR
    \Input{$\boldsymbol{A} \in \mathbb{R}^{m \times d}$, $G_{\boldsymbol{\theta}}: \mathbb{R}^{v} \mapsto \mathbb{R}^{d}$, $(\tau, \eta, \lambda) \in \mathbb{R}_{+}^{3}$, $T \in \mathbb{N}$}
    \Initialize{$\boldsymbol{\theta}(0) = \boldsymbol{\theta}$}
        \For{$k = 0$ to $K-1$}
    	\For{$i = 1$ to $n$}
    	\State sample $\boldsymbol{x}_{tr,i}$ from $p(\boldsymbol{x})$  
    	\State measure $\boldsymbol{y}_{tr,i}=\boldsymbol{A} \boldsymbol{x}_{tr,i}$
    	\State sample $\boldsymbol{z}_{i,0}$ from $p_{\boldsymbol{z}}(\boldsymbol{z})$  
    	\For{$t = 0$ to $T-1$}
    	\State $\boldsymbol{z}_{i,t+1} = \boldsymbol{z}_{i,t} - \tau \frac{\partial}{\partial \boldsymbol{z}}\left\| \boldsymbol{y}_{tr,i} - \boldsymbol{A}G_{\boldsymbol{\theta}}(\boldsymbol{z}, \begin{color}{blue}\boldsymbol{y}_{tr,i}\end{color}) \right\|^2\Big|_{\boldsymbol{z}=\boldsymbol{z}_{i,t}}$
    	\EndFor
    	\EndFor
    	\Comment{generate $n$ training samples}
    	\vskip +1pt
    	\State $M({\boldsymbol{\theta}}) = \frac{1}{n} \sum\limits_{i=1}^{n} \left\|\boldsymbol{y}_{tr,i}- \boldsymbol{A}G_{\boldsymbol{\theta}}(\boldsymbol{z}_{i,T}, \begin{color}{blue}\boldsymbol{y}_{tr,i}\end{color})\right\|^2$ 
    	\State $R({\boldsymbol{\theta}}) = \frac{1}{n}\sum\limits_{i=1}^n \mathbb{E}_{\boldsymbol{x}_1,\boldsymbol{x}_2 \in \{\boldsymbol{x}_{tr,i},G_{\boldsymbol{\theta}}(\boldsymbol{z}_{i,0}, \begin{color}{blue}\boldsymbol{y}_{tr,i}\end{color}),G_{\boldsymbol{\theta}}(\boldsymbol{z}_{i,T}, \begin{color}{blue}\boldsymbol{y}_{tr,i}\end{color})\}}\Big[\big(\left\|  \boldsymbol{A}(\boldsymbol{x}_1-\boldsymbol{x}_2) \right\| - \left\| \boldsymbol{x}_1-\boldsymbol{x}_2\right\|\big)^2\Big]$
    	\State $\boldsymbol{\theta}(k+1) = \boldsymbol{\theta}(k) - \eta\frac{\partial}{\partial \boldsymbol{\theta}}\Big(M({\boldsymbol{\theta}})+\lambda R({\boldsymbol{\theta}})  \Big) \Big|_{\boldsymbol{\theta}=\boldsymbol{\theta}(k)}$ \Comment{optimize the network parameters}
    	\EndFor
        \Output{the trained parameters ${\boldsymbol{\theta}({K})}$ of $G$}
  \end{algorithmic}
  \algrule
    \begin{algorithmic}[H]
    \TE 
    \Input{$\boldsymbol{y}_{te} \in \mathbb{R}^{m}, \boldsymbol{A} \in \mathbb{R}^{m \times d}$, $G_{{\boldsymbol{\theta}}(K)}: \mathbb{R}^{v} \mapsto \mathbb{R}^{d}$, $\tau  \in \mathbb{R}_{+}$, $T\in \mathbb{N}$}
    \Initialize{sample $\boldsymbol{z}_0$ from $p_{\boldsymbol{z}}(\boldsymbol{z})$}
	\For{$t = 0$ to $T-1$}
	\State $\boldsymbol{z}_{t+1} = \boldsymbol{z}_{t} - \tau \frac{\partial}{\partial \boldsymbol{z}}\left\| \boldsymbol{y}_{te} - \boldsymbol{A}G_{\boldsymbol{\theta}({K})}(\boldsymbol{z}, \begin{color}{blue}\boldsymbol{y}_{te}\end{color}) \right\|^2\Big|_{\boldsymbol{z}=\boldsymbol{z}_t}$
	\EndFor
    \Output{the signal estimate $\hat{\boldsymbol{x}} = G_{\boldsymbol{\theta}({K})}(\boldsymbol{z}_T, \begin{color}{blue}\boldsymbol{y}_{te}\end{color})$}	
  \end{algorithmic}
  \label{alg_dcs}
\end{algorithm}

\begin{algorithm}
  \caption{SparseGen (without blue notes) and SparseGen-IM in the case of using DCGAN training objective}
    \begin{algorithmic}[H]
    \TR: 
    \Input{$\boldsymbol{A} \in \mathbb{R}^{m \times d}$, $G_{\boldsymbol{\theta}}: \mathbb{R}^{v} \mapsto \mathbb{R}^{d}$, $ D_{\boldsymbol{\phi}}: \mathbb{R}^{d} \mapsto [0, 1]$}
	\For{$i = 1$ to $n$}
	\State sample $\boldsymbol{x}_{tr,i}$ from $p(\boldsymbol{x})$  
	\State measure $\boldsymbol{y}_{tr,i}=\boldsymbol{A} \boldsymbol{x}_{tr,i}$
	\State sample $\boldsymbol{z}_{i}$ from $p_{\boldsymbol{z}}(\boldsymbol{z})$  
	\EndFor
	\Comment{generate $n$ training samples}
	\vskip +1pt
	\State $({\boldsymbol{\theta}}^*,{\boldsymbol{\phi}}^*)=\underset{\boldsymbol{\theta}}{\argmin}\, \underset{\boldsymbol{\phi}}{\argmax}\,\frac{1}{n}\sum\limits_{i=1}^n [\ln \, D_{\boldsymbol{\phi}}(\boldsymbol{x}_{tr,i}, \begin{color}{blue}\boldsymbol{y}_{tr,i}\end{color}) + \ln \, (1-D_{\boldsymbol{\phi}}(G_{\boldsymbol{\theta}}(\boldsymbol{z}_i, \begin{color}{blue}\boldsymbol{y}_{tr,i}\end{color}), \begin{color}{blue}\boldsymbol{y}_{tr,i}\end{color}))]$ 
    \Output{the trained parameters ${\boldsymbol{\theta}}^*$ of $G$}
    \end{algorithmic}
  \algrule
    \begin{algorithmic}[H]
    \TE
    \Input{$\boldsymbol{y}_{te} \in \mathbb{R}^{m}, \boldsymbol{A} \in \mathbb{R}^{m \times d}$, $G_{\boldsymbol{\theta}^*}: \mathbb{R}^{v} \mapsto \mathbb{R}^{d}$, $\tau  \in \mathbb{R}_{+}$, $(L, T) \in \mathbb{N}^2$ where $L < T$}
    \Initialize{sample $\boldsymbol{z}_0$ from $p_{\boldsymbol{z}}(\boldsymbol{z})$, $\boldsymbol{\nu}_0 = \boldsymbol{0}$}
	\For{$t = 0$ to $T-1$}
	\If{$t < L$}
	\State $\boldsymbol{z}_{t+1} = \boldsymbol{z}_{t} - \tau \frac{\partial}{\partial \boldsymbol{z}}\Big(\left\| \boldsymbol{A}(G_{{\boldsymbol{\theta}}^*}(\boldsymbol{z}, \textcolor{blue}{\boldsymbol{y}_{te}}) + \boldsymbol{\nu}_t)-\boldsymbol{y}_{te} \right\|^2 + \lambda \left\| \boldsymbol{B}\boldsymbol{\nu}_t\right\|_1\Big) \Big|_{\boldsymbol{z}=\boldsymbol{z}_t}$
	\State $\boldsymbol{\nu}_{t+1} = \boldsymbol{\nu}_t$
	\Else
	\State $\boldsymbol{\mathcal{T}} = (\boldsymbol{z}, \boldsymbol{\nu})$
	\State $(\boldsymbol{z}_{t+1}, \boldsymbol{\nu}_{t+1}) = (\boldsymbol{z}_{t}, \boldsymbol{\nu}_{t}) - \tau \frac{\partial}{\partial \boldsymbol{\mathcal{T}}}\big(\left\| \boldsymbol{A}(G_{{\boldsymbol{\theta}}^*}(\boldsymbol{z}, \textcolor{blue}{\boldsymbol{y}_{te}}) + \boldsymbol{\nu})-\boldsymbol{y}_{te} \right\|^2 + \lambda \left\| \boldsymbol{B}\boldsymbol{\nu}\right\|_1\big) \Big|_{\boldsymbol{\mathcal{T}} = (\boldsymbol{z}_t, \boldsymbol{\nu}_t)}$
	\EndIf
	\EndFor
    \Output{the signal estimate $\hat{\boldsymbol{x}} = G_{\boldsymbol{\theta}^*}(\boldsymbol{z}_{T}, \begin{color}{blue}\boldsymbol{y}_{te}\end{color})+ \boldsymbol{\nu}_T$}
  \end{algorithmic}
  \label{alg_sparsegen}
\end{algorithm}

\clearpage
\section{Experimental Details}

When adding the measurement vector $\boldsymbol{y}$ into the DCGAN architecture, we emulate a conventional architecture suggested in \citet{reed2016textgan} since $\boldsymbol{y}$ can be considered as the text description embedding. In the generator $G_{\boldsymbol{\theta}}$, $\boldsymbol{y}$ is concatenated to the noise vector $\boldsymbol{z}$. In the discriminator $D_{\boldsymbol{\phi}}$, $\boldsymbol{y}$ is duplicated spatially and concatenated to the $4 \times 4$ sized image feature maps in a channel-wise manner. To reduce the number of channels to the original number of ones, the concatenated feature maps pass through a $1 \times 1$ convolution followed by a batch normalization and the rectified linear unit function. 

When putting $\boldsymbol{y}$ in the BEGAN architecture, $\boldsymbol{y}$ is solely concatenated to latent variables in the generator and the decoder of the discriminator, which is far simpler than the DCGAN architecture supplemented with $\boldsymbol{y}$.

In Figure \ref{fig:im_effect}, we employ CSGM and CSGM-IM using DCGAN trained on CelebA dataset. Red/blue/green-colored `x' indicates the first/second/third sample in the test set of CelebA. We sample $50$ samples from $G_{\boldsymbol{\theta}^*}(\boldsymbol{z}, \boldsymbol{y}_{te})$ (CSGM-IM) per test sample and $150$ samples from $G_{\boldsymbol{\theta}^*}(\boldsymbol{z})$ (CSGM). In Figure \ref{fig:z_effect}, each curve represents the average reconstruction error over 64 test samples per iteration, where the experimental setting of Figure \ref{fig:z_effect} is the same as that of Figure \ref{fig:im_effect} except the dimension of $z$: the dimension of $z$ in CSGM-IM is equal to $m$ for better visualization of Figure \ref{before vs. after}, and that of $z$ in CSGM is set to $2m$.

For experiments on PGDGAN and PGDGAN-IM in Section \ref{exp}, we follow the same experimental setting as \citet{shah2018solving} such as $\alpha = 0.5$, $\tau = 0.1$, $T = 10$, and $K = 100$. 

When conducting experiments on DCS and DCS-IM in Section \ref{exp}, a sensing matrix $A$ is not learned and the learning rate in the latent optimization ($\tau$ in Algorithm \ref{alg_dcs}) is fixed as $0.01$ for brevity. When implementing DCS, $\lambda$ in Algorithm \ref{alg_dcs} is set to $1.0$ like \citet{wu2019deep} for $m \le 1000$, but $0.001$ for $m > 1000$ 
because a large value of $R(\boldsymbol{\theta})$ in Algorithm \ref{alg_dcs} hinders DCS from performing well. 
Furthermore, $T = 5$ for $m = 5000$ in DCS, whereas we follow the same hyperparameters ($\lambda = 1.0$, $T = 3$) as \citet{wu2019deep} for all $m$ in DCS-IM. 

When running experiments on SparseGen and SparseGen-IM in Section \ref{exp}, $\lambda$ in \eqref{sparsegen} and \eqref{sparsegen-im} (or Algorithm \ref{alg_sparsegen}) is chosen among $\{0.1, 0.5, 1.0\}$ for each $m$.
Other than $\lambda$, we make use of the identical hyperparameters ($L = 250$, $T = 500$ in Algorithm \ref{alg_sparsegen}) as introduced in \citet{dhar2018modeling}.
Note that we solely consider a transform matrix $\boldsymbol{B}$ in Algorithm \ref{alg_sparsegen} as the wavelet basis given that \citet{dhar2018modeling} recommend the wavelet basis rather than the discrete cosine transform. 

The fastMRI dataset consists of $34,732$ training slices, $7,135$ validation slices, and $3,903$ test slices. Owing to the low quality of test slices, we regard the validation slices as the test set and thus utilize $64$ random images in the validation slices for inference. As the values of the original slices are complex, we preprocess the data as follows: (i) apply two-dimensional fast Fourier transform to each slice, (ii) crop it at the center to the size $256 \times 256$, (iii) take the absolute value of a complex-valued slice, (iv) subtract $0.5$ and divide by $0.5$, and (v) downsample it to the size $128 \times 128$. Any data augmentation is not used to show that IM can encourage previous methods to perform well even on a small number of data. Same as Section \ref{exp}, each entry of $\boldsymbol{A}$ is also sampled from $\mathcal{N}(0, 1/m)$, and the reconstruction error is used for evaluation.

\clearpage
\section{Additional Experimental Results}
\begin{table}[H]
    \parbox{.49\linewidth}{
    \vskip -10pt
	\caption{\footnotesize Reconstruction error per pixel with $95\%$ confidence interval of $5$ trials for Section \ref{subsec:csgm} using DCGAN.}
	\centering
 	{\footnotesize
	\begin{tabular}{lcc}
		\toprule
		$m$ & CSGM & CSGM-IM \\
		\midrule
		$20$ & $0.3038 \pm 0.0678$ & $\mathbf{0.2088 \pm 0.0106}$ \\
		$50$ & $0.1859 \pm 0.0096$ & $\mathbf{0.1131 \pm 0.0077}$ \\
		$100$ & $0.1043 \pm 0.0116$ & $\mathbf{0.0720 \pm 0.0069}$ \\
		$200$ & $0.0624 \pm 0.0069$ & $\mathbf{0.0464 \pm 0.0039}$ \\
		$500$ & $0.0392 \pm 0.0039$ & $\mathbf{0.0286 \pm 0.0031}$ \\
		$1000$ & $0.0332 \pm 0.0031$ & $\mathbf{0.0217 \pm 0.0025}$ \\
		$2500$ & $0.0298 \pm 0.0036$ & $\mathbf{0.0185 \pm 0.0017}$ \\
		$5000$ & $0.0285 \pm 0.0031$ & $\mathbf{0.0183 \pm 0.0018}$ \\ 
		\bottomrule
	\end{tabular}
	} 
	\label{tab:CSGM_DCGAN_5trials}
    }
    \parbox{.49\linewidth}{
    \vskip -10pt
	\caption{\footnotesize Reconstruction error per pixel with $95\%$ confidence interval of $5$ trials for Section \ref{subsec:csgm} using BEGAN.}
	\centering
 	{\footnotesize
	\begin{tabular}{lcc}
		\toprule
		$m$ & CSGM & CSGM-IM \\
		\midrule
		$20$ & $0.2125 \pm 0.0314$ & $\mathbf{0.1855 \pm 0.0166}$ \\
		$50$ & $0.1309 \pm 0.0218$ & $\mathbf{0.1008 \pm 0.0114}$ \\
		$100$ & $0.0990 \pm 0.0127$ & $\mathbf{0.0580 \pm 0.0067}$ \\
		$200$ & $0.0805 \pm 0.0096$ & $\mathbf{0.0331 \pm 0.0038}$ \\
		$500$ & $0.0711 \pm 0.0111$ & $\mathbf{0.0147 \pm 0.0022}$ \\
		$1000$ & $0.0680 \pm 0.0095$ & $\mathbf{0.0075 \pm 0.0012}$ \\
		$2500$ & $0.0669 \pm 0.0095$ & $\mathbf{0.0031 \pm 0.0005}$ \\
		$5000$ & $0.0660 \pm 0.0109$ & $\mathbf{0.0020 \pm 0.0003}$ \\
		\bottomrule
	\end{tabular}
	} 
	\label{tab:CSGM_BEGAN_5trials}
    }
\end{table}

\begin{table}[H]
    \parbox{.49\linewidth}{
    \vskip -2pt
	\caption{\footnotesize Reconstruction error per pixel with $95\%$ confidence interval of $5$ trials for Section \ref{subsec:pgd} using DCGAN.}
	\centering
 	\footnotesize
 	{
	\begin{tabular}{lcc}
		\toprule
		$m$ & PGDGAN & PGDGAN-IM \\
		\midrule
		$20$ & $0.6298 \pm 0.0757$ & $\mathbf{0.4415 \pm 0.0205}$ \\
		$50$ & $0.2730 \pm 0.0353$ & $\mathbf{0.1626 \pm 0.0105}$ \\
		$100$ & $0.1281 \pm 0.0101$ & $\mathbf{0.0942 \pm 0.0067}$ \\
		$200$ & $0.0793 \pm 0.0065$ & $\mathbf{0.0539 \pm 0.0046}$ \\
		$500$ & $0.0489 \pm 0.0063$ & $\mathbf{0.0311 \pm 0.0030}$ \\
		$1000$ & $0.0383 \pm 0.0037$ & $\mathbf{0.0234 \pm 0.0025}$ \\
		$2500$ & $0.0337 \pm 0.0035$ & $\mathbf{0.0195 \pm 0.0017}$ \\
		$5000$ & $0.0322 \pm 0.0035$ & $\mathbf{0.0194 \pm 0.0018}$ \\ 
		\bottomrule
	\end{tabular}
	}
	\label{tab:PGD_DCGAN_5trials}
    }
    \parbox{.49\linewidth}{
    \vskip -2pt
	\caption{\footnotesize Reconstruction error per pixel with $95\%$ confidence interval of $5$ trials for Section \ref{subsec:pgd} using DCGAN.}
	\centering
 	{\footnotesize
	\begin{tabular}{lcc}
		\toprule
		$m$ & SPGDGAN & SPGDGAN-IM \\
		\midrule
		$20$ & $0.6243 \pm 0.1100$ & $\mathbf{0.4250 \pm 0.0211}$ \\
		$50$ & $0.2644 \pm 0.0161$ & $\mathbf{0.1552 \pm 0.0069}$ \\
		$100$ & $0.1245 \pm 0.0103$ & $\mathbf{0.0899 \pm 0.0076}$ \\
		$200$ & $0.0778 \pm 0.0078$ & $\mathbf{0.0521 \pm 0.0044}$ \\
		$500$ & $0.0482 \pm 0.0054$ & $\mathbf{0.0308 \pm 0.0031}$ \\
		$1000$ & $0.0385 \pm 0.0040$ & $\mathbf{0.0235 \pm 0.0024}$ \\ 
		$2500$ & $0.0333 \pm 0.0029$ & $\mathbf{0.0198 \pm 0.0017}$ \\
		$5000$ & $0.0325 \pm 0.0044$ & $\mathbf{0.0196 \pm 0.0018}$ \\
		\bottomrule
	\end{tabular}
	}
	\label{tab:SPGD_DCGAN_5trials}
	}
\end{table}

\begin{table}[H]
    \parbox{.49\linewidth}{
    \vskip -2pt
    \caption{\footnotesize Reconstruction error per pixel with $95\%$ confidence interval of $5$ trials for Section \ref{subsec:dcs} only using the generator of DCGAN.}
	\centering
 	{\footnotesize
	\begin{tabular}{lcc}
		\toprule
		$m$ & DCS & DCS-IM \\
		\midrule
		$20$ & $0.2460 \pm 0.0052$ & $\mathbf{0.1095 \pm 0.0081}$ \\
		$50$ & $0.2190 \pm 0.0175$ & $\mathbf{0.0694 \pm 0.0067}$ \\
		$100$ & $0.1587 \pm 0.0072$ & $\mathbf{0.0489 \pm 0.0053}$ \\
		$200$ & $0.1789 \pm 0.0077$ & $\mathbf{0.0329 \pm 0.0036}$ \\
		$500$ & $0.1136 \pm 0.0071$ & $\mathbf{0.0173 \pm 0.0023}$ \\
		$1000$ & $0.0874 \pm 0.0032$ & $\mathbf{0.0099 \pm 0.0014}$ \\ 
		$2500$ & $0.0854 \pm 0.0055$ & $\mathbf{0.0036 \pm 0.0005}$ \\
		$5000$ & $0.0817 \pm 0.0024$ & $\mathbf{0.0024 \pm 0.0003}$ \\
		\bottomrule
	\end{tabular}
	}
	\label{tab:DCS_DCGAN_5trials}
    }
    \parbox{.49\linewidth}{
    \vskip -2pt
		\caption{\footnotesize Reconstruction error per pixel with $95\%$ confidence interval of $5$ trials for Section \ref{subsec:sparse} using DCGAN.}
	\centering
 	\footnotesize
 	{
	\begin{tabular}{lcc}
		\toprule
		$m$ & SparseGen & SparseGen-IM \\
		\midrule
		$20$ & $0.3743 \pm 0.0598$ & $\mathbf{0.2242 \pm 0.0218}$ \\
		$50$ & $0.2125 \pm 0.0390$ & $\mathbf{0.1121 \pm 0.0103}$ \\
		$100$ & $0.1184 \pm 0.0180$ & $\mathbf{0.0720 \pm 0.0054}$ \\
		$200$ & $0.0691 \pm 0.0158$ & $\mathbf{0.0470 \pm 0.0037}$ \\
		$500$ & $0.0378 \pm 0.0050$ & $\mathbf{0.0283 \pm 0.0031}$ \\
		$1000$ & $0.0302 \pm 0.0040$ & $\mathbf{0.0213 \pm 0.0024}$ \\
		$2500$ & $0.0248 \pm 0.0028$ & $\mathbf{0.0178 \pm 0.0016}$ \\
		$5000$ & $0.0238 \pm 0.0025$ & $\mathbf{0.0174 \pm 0.0017}$ \\ 
		\bottomrule
	\end{tabular}
	}
	\label{tab:Sparse_DCGAN_5trials}
	}
\end{table}

\end{document}